\pdfoutput=1 % tell arXiv to use PDFLaTeX

\documentclass[journal]{IEEEtran}

\usepackage[T1]{fontenc} % output - specifies encoding used in fonts; needs full LaTeX distribution to produce good-looking output
\usepackage[utf8]{inputenc} % input - type accented characters directly from keyboard
\usepackage[english]{babel} % internationalization - hyphenation, typographic rules for one or more languages

\usepackage[nounderscore]{syntax} % definition of the context-free grammar

\usepackage[
  backend=biber, % bibliography engine
  style=ieee, % bibliography style .bbx and citation style .cbx
  hyperref, % make citations and references clickable - requires hyperref pkg
  maxbibnames=99, % 99 = display all authors of multi-author articles
  doi=false, % do not display DOI
  isbn=false, % do not display ISBN
]{biblatex} % note: incompatible with ucs (-> utf8x), natbib
\renewcommand{\bibfont}{\footnotesize}
\usepackage[
  autostyle % adapt citation style to current document language
]{csquotes} % Context Sensitive Quotations; provides biblatex \enquote{}

\usepackage{hyperref}
\let\oldFootnote\footnote
\newcommand\nextToken\relax
\renewcommand\footnote[1]{%
    \oldFootnote{#1}\futurelet\nextToken\isFootnote}
\newcommand\isFootnote{%
    \ifx\footnote\nextToken\textsuperscript{,}\fi}

\usepackage[nolist]{acronym}
\usepackage{amsmath}
\usepackage{amssymb}
\usepackage{bm} % bold math
\usepackage{booktabs}
\usepackage{graphicx}
\newcommand{\myWidth}{0.16} % used in various figures involving photos
\usepackage[multiple]{footmisc} % multiple footnote at the same point
\usepackage{siunitx}
\usepackage{subfig}
\usepackage{tabularx}
\usepackage{url}

\usepackage{tikz}
\usetikzlibrary{matrix,positioning}
\usetikzlibrary{decorations.text}
\usetikzlibrary{shapes.geometric}
\usetikzlibrary{fit}
\tikzstyle{circle}=[shape=circle,minimum size=0.7cm,very thick]
\tikzstyle{every path}=[very thick]
\usetikzlibrary{arrows}

\usepackage[update,prepend]{epstopdf}
\graphicspath{{figures/}{bios/}}

\newcommand{\actionperception}{action--perception}
\newcommand{\AffWords}{Affordance--Words}
\newcommand{\FB}{Forward--Backward}

\newcommand{\hh}{human--human}
\newcommand{\hr}{human--robot}
\newcommand{\hri}{\hr{} interaction}
\newcommand{\ObjAct}{Object--Action}
\newcommand{\objecthand}{object--hand}

\newcommand{\selfother}{self--other}
\newcommand{\wordmeaning}{word--meaning}

\newcommand{\phmm}{\ensuremath{P_{\text{HMM}}}}
\newcommand{\pbn}{\ensuremath{P_{\text{BN}}}}
\newcommand{\pcomb}{\ensuremath{P_\text{comb}}}
\newcommand{\xinf}{\ensuremath{X_\text{inf}}}
\newcommand{\xobs}{\ensuremath{X_\text{obs}}}
\newcommand{\xlat}{\ensuremath{X_\text{lat}}}

\newcommand{\given}{\ensuremath{\mid}}

\begin{acronym}
\acro{AI}{Artificial Intelligence}
\acro{BN}{Bayesian Network}
\acro{CFG}{context-free grammar}
\acro{HMM}{Hidden Markov Model}
\acro{OAC}{\ObjAct{} Complex}
\acrodefplural{OAC}{\ObjAct{} Complexes}
\acro{PDF}{Probability Density Function}
\end{acronym}

\begin{document}

\title{Beyond the Self: Using Grounded Affordances to Interpret and Describe Others' Actions}

\author{Giovanni~Saponaro,~\IEEEmembership{Student Member,~IEEE,}
        Lorenzo~Jamone,~\IEEEmembership{Member,~IEEE,}
        Alexandre~Bernardino,~\IEEEmembership{Senior~Member,~IEEE}
        Giampiero~Salvi,~\IEEEmembership{Member,~IEEE}
        \thanks{
          Manuscript received November 15, 2017; revised September 6, 2018; accepted November 14, 2018. This research was supported by the FCT projects~UID/EEA/50009/2013, AHA~CMUP-ERI/HCI/0046/2013 and by the CHIST-ERA project IGLU.}
\thanks{G.~Saponaro and A.~Bernardino are with the
Institute for Systems and Robotics, Instituto Superior Técnico,
Universidade de Lisboa, Lisbon, Portugal, e-mail: \{gsaponaro,alex\}@isr.tecnico.ulisboa.pt.}
\thanks{L.~Jamone is with ARQ~(Advanced Robotics at Queen Mary), School of Electronic Engineering and Computer Science, Queen Mary University of London, United Kingdom
and with the
Institute for Systems and Robotics, Instituto Superior Técnico, Universidade de Lisboa, Lisbon, Portugal,
e-mail: l.jamone@qmul.ac.uk.}
\thanks{G.~Salvi is with KTH Royal Institute of Technology, Stockholm, Sweden,
  e-mail: giampi@kth.se.}
}

% make the title area
\maketitle
\IEEEpeerreviewmaketitle

%%%%%%%%%%%%%%%%%%%%%%%%%%%%%%%%%%%%%%%%%%%%%%%%%%%%%%%%%%%%%%%%%%%%%%%%%%%%%%%%
\begin{abstract}
  %!TEX encoding = UTF-8 Unicode
We propose a developmental approach that allows a robot to interpret and describe the actions of human agents by reusing previous experience.
The robot first learns the association between words and object affordances by manipulating the objects in its environment.
It then uses this information to learn a mapping between its own actions and those performed by a human in a shared environment.
It finally fuses the information from these two models to interpret and describe human actions in light of its own experience.
In our experiments, we show that the model can be used flexibly to do inference on different aspects of the scene.
We can predict the effects of an action on the basis of object properties.
We can revise the belief that a certain action occurred, given the observed effects of the human action.
In an early action recognition fashion, we can anticipate the effects when the action has only been partially observed.
By estimating the probability of words given the evidence and feeding them into a pre-defined grammar, we can generate relevant descriptions of the scene.
We believe that this is a step towards providing robots with the fundamental skills to engage in social collaboration with humans.

\end{abstract}


%%%%%%%%%%%%%%%%%%%%%%%%%%%%%%%%%%%%%%%%%%%%%%%%%%%%%%%%%%%%%%%%%%%%%%%%%%%%%%%%
%!TEX encoding = UTF-8 Unicode

% this figure should come before the proposed approach section.
% It might happen automatically if we shorten the introduction.
% Otherwise move back in the files.
\begin{figure*}
  \centering
  \subfloat[][Grasp action: moving the hand towards an object vertically, then grasping and lifting it.]{
    \resizebox{\linewidth}{!}{
      \includegraphics{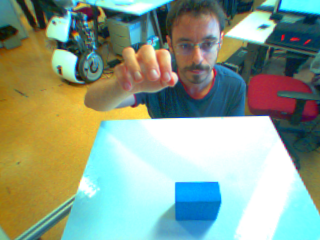}
      \includegraphics{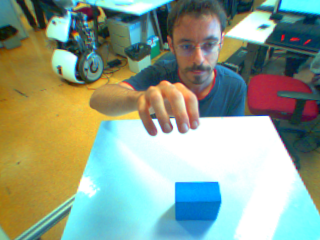}
      \includegraphics{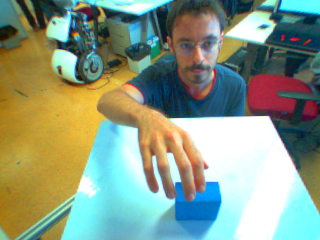}
      \includegraphics{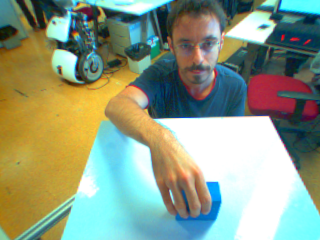}
      \includegraphics{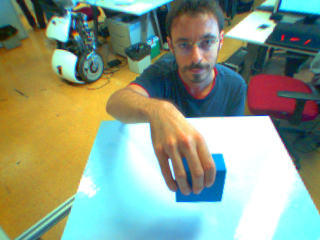}
      \includegraphics{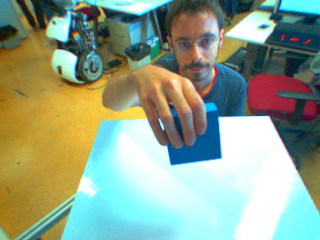}
    } % end resizebox
    \label{fig:action_examples:grasp}
  } % end subfloat

  \subfloat[][Tap action: moving the hand towards an object laterally then touching it, causing a motion effect.]{
    \resizebox{\linewidth}{!}{
      \includegraphics{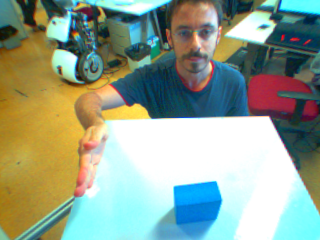}
      \includegraphics{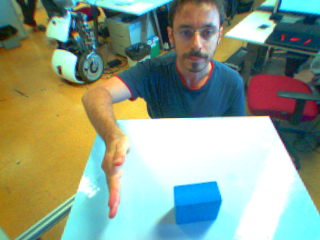}
      \includegraphics{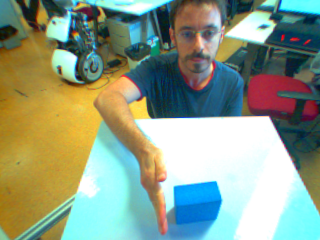}
      \includegraphics{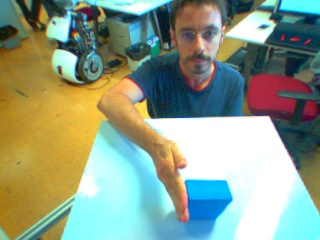}
      \includegraphics{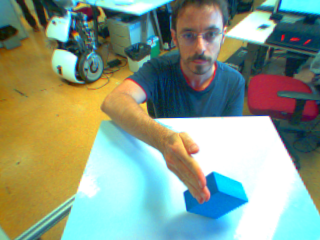}
      \includegraphics{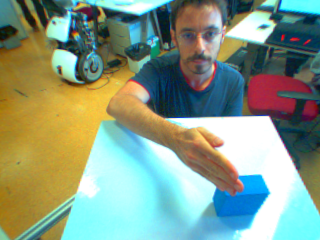}
    } % end resizebox
    \label{fig:action_examples:tap}
  } % end subfloat

  \subfloat[][Touch action: moving the hand towards an object vertically, touching it~(without grasping), then retracting the hand.]{
    \resizebox{\linewidth}{!}{
      \includegraphics{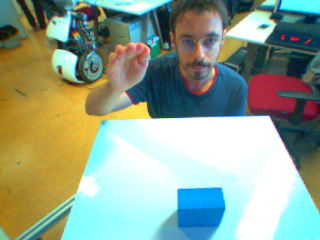}
      \includegraphics{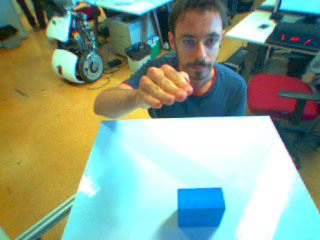}
      \includegraphics{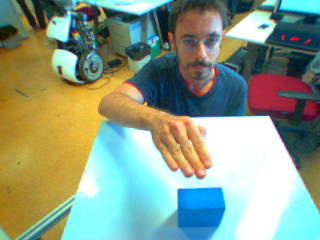}
      \includegraphics{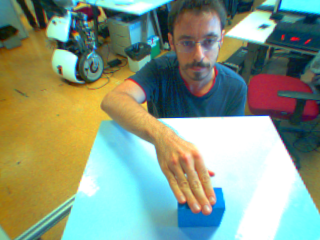}
      \includegraphics{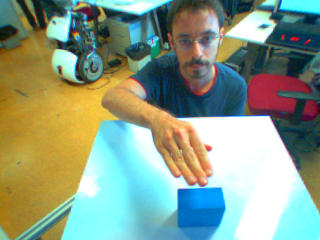}
      \includegraphics{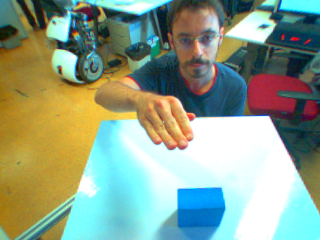}
    } % end resizebox
    \label{fig:action_examples:touch}
  } % end subfloat
  \caption{Examples of human actions from the point of view of the robot.}
  \label{fig:action_examples}
\end{figure*}

\section{Introduction}
\label{sec:intro}

\IEEEPARstart{C}{ooperation}, or the ability of working successfully in groups, is a tenet of human society~\cite{turner:1975}.
This skill is acquired by human children incrementally, around the second year of life, as they develop the ability to coordinate themselves with peers or adult caregivers in shared problem-solving activities and social games~\cite{brownell:2006:childdev}.
This is achieved not only by mere behavioral coordination, but also by employing communicative strategies~\cite{melis:2010:rstb} and by continuously observing partners' actions~\cite{ramnani:2004:natureneuro}.
Loosely inspired by these observations, this article presents and evaluates a cognitive system for robots which permits reasoning over subsequent phases:
first about self-learned knowledge~(about affordances and language-based descriptions of objects),
and then about others' actions.

Even though social robots\footnote{A social robots is ``[a robot that is] able to communicate and interact with us, understand and even relate to us, in a personal way. [It] should be able to understand us and itself in social terms''~\cite{breazeal:2002:dsr}.} are becoming common in domestic and public environments, \hr{} teams still lag behind \hh{} teams in terms of effectiveness.
For robots, interpreting the actions of others and learning to describe them verbally~(for effective cooperation) is challenging.
The reason is that we cannot possibly model all the imaginable physical, verbal and non-verbal~(e.g., gestures) cues that can take place during \hri, due to the richness of language and the high variability of the real world outside of structured research laboratories and factories.
Hence, it is necessary to have robots that \emph{learn} world elements and properties of language~\cite{iwahashi:2007:hri}, and the ability to link these verbal elements with other skills, such as other perceptual modalities~(e.g., vision of objects and other agents) and manipulation abilities~(e.g., grasping objects and placing them in order to achieve a goal)~\cite{steels:2003:trendscogsci}.

Our work builds upon the intuition that a robot can generalize its previously-acquired knowledge of the world~(e.g., motor actions, objects properties, physical effects, verbal descriptions) to those situations where it observes a human agent performing familiar actions in a shared \hr{} scenario.
We follow the developmental robotics perspective~\cite{lungarella:2003:devrobsurvey,cangelosi:2015:devrobbook},
which takes inspiration from the progressive learning phenomena observed in children's mental development~(e.g., the understanding of language, the acquisition of manipulation skills, the comprehension of others' actions), and investigates how to model the evolution and acquisition of these increasingly complex cognitive processes in artificial autonomous systems.

In particular, we are inspired by the possible existence of a shared representation for self-related and others-related knowledge in the human brain~\cite{gallagher:1996:earliest,rizzolatti:2001:nrn,decety:2003:sharedrep}, and we look at the developmental stages in which human children have consolidated an idea of \selfother{} distinction~\cite{symons:2004:mental} and start to reason about the external world also in allocentric terms~\cite{ribordy:2013:development}, in addition to the ego-centric ones, and could therefore possibly begin to use knowledge about the \emph{self} to infer about \emph{others}.

Extending on our recent work~\cite{saponaro:2017:glu}, in this article we combine robot ego-centric learning about language and object affordances~\cite{salvi:2012:smcb} with the observation of external agents through gesture recognition~\cite{saponaro:2013:crhri}.
Our novel contributions are:
(i)~a probabilistic method to fuse self-learned knowledge of language and object affordances, with socially aware information of others' physical actions~(in the form of uncertain soft evidence);
(ii)~experimental findings showing the reasoning power of our combined system, which is able to make inferences and predictions over affordances and words; and
(iii)~the possibility of generating verbal descriptions from the estimated word probabilities and a pre-defined grammar, with emergence of non-trivial language properties such as congruent/incongruent conjunctions, synonyms between two consecutive sentences speaking about the same concepts.
Furthermore, we make our human action data and probabilistic reasoning code publicly available\footnote{\url{https://github.com/giampierosalvi/AffordancesAndSpeech}: the code from \cite{salvi:2012:smcb} has been extended to support the experiments in this study.}\footnote{\url{https://github.com/gsaponaro/tcds-gestures}: code from this paper.} in the interest of reproducibility.

This article is structured as follows.
In Sec.~\ref{sec:related_work} we briefly overview the literature on the interpretation and verbal description of others in different disciplines,
in Sec.~\ref{sec:method} we present our proposed method and its components,
in Sec.~\ref{sec:experimental_settings} we provide details and assumptions of the approach,
Sec.~\ref{sec:results} illustrates our results, and
in Sec.~\ref{sec:conclusions} we draw our concluding remarks.

%%%%%%%%%%%%%%%%%%%%%%%%%%%%%%%%%%%%%%%%%%%%%%%%%%%%%%%%%%%%%%%%%%%%%%%%%%%%%%%%
%!TEX encoding = UTF-8 Unicode

\section{Related Work}
\label{sec:related_work}

Human cooperation is a phenomenon that we often take for granted~(at least in adults), possibly because it is widespread and intimately embedded into human societies.
However, this non-trivial skill is greatly facilitated, and influenced, by human language~\cite{mueller:2000:psych}.
For instance, educational research has shown that, when language is used as a cultural tool for intellectual tasks in preteen students, discursive interaction enables collective thinking to become more effective, also fostering individual reasoning and faster learning~\cite{rojas:2003:ijer}.

The ability to understand and interpret our peers has also been studied in neuroscience and psychology, focusing on internal simulations and re-enactments of previous experiences~\cite{schillaci:2012:hbu,billing:2016:frobt}, or on visuomotor neurons~\cite{rizzolatti:2001:nrn}, i.e., neurons that are activated by visual stimuli.
Mirror neurons respond to action and object interaction, both when the agent acts and when it observes the same action performed by others, hence the name ``mirror''.
They are based on the principle that perceptual input can be linked with the human action system for predicting future outcomes of actions, i.e., the effect of actions, particularly when the person possesses concrete prior personal experience of the actions being observed in others~\cite{aglioti:2008:basketball,knoblich:2001:psychsci}.

In applying the mirror neuron theory in robotics, as we and others do~\cite{gazzola:2007:neuroimage,lopes:2009:ab}, an agent can first acquire knowledge by sensing and self-exploring its surrounding environment.
Afterwards, it can employ that learned knowledge to novel observations of another agent~(e.g., a human person) who performs similar physical actions to the ones executed during prior training.
In particular, when the two interacting agents are a caregiver and an infant, the mechanism is called \emph{parental scaffolding}, having been implemented on robots too~\cite{ugur:2015:robotica,ugur:2015:tamd}.
These works tackle the so-called correspondence problem~\cite{nehaniv:2002:correspondence}, in our case in a simple collaboration scenario, assuming that the two agents are capable of applying actions to objects leading to similar effects, enabling the transfer, and that they operate on a shared space~(i.e., a table accessible by both agents' arms).
The morphology and the motor realization of the actions can be different between the two agents.

Some authors have studied the ability to interpret other agents under the deep learning paradigm.
In~\cite{kim:2017:nn}, a recurrent neural network is proposed to have an artificial simulated agent infer human intention~(as output) from joint input information about objects, their potential affordances or opportunities, and human actions, employing different time scales for different actions.
However, in that work a virtual simulation able to produce large quantities of data was used.
This is both unrealistic when trying to explain human cognition, and limited, because a simulator cannot model all the physical events and the unpredictability of the real world.
In contrast, we use real, noisy data acquired from robots and sensors to validate our model.
In addition, deep neural networks trained with large amounts of data can be difficult to inspect in their inner layers and activations~\cite{szegedy:2014:intriguing}, whereas our Bayesian model is focused on exhibiting emerging patterns of causality, choices, explanations from relatively few data points.

DeepMind and Google published a method~\cite{santoro:2017:relational_reasoning} to perform relational reasoning on images, i.e., a system that learns to reflect about entities and their mutual relations, with the ability of providing answers to questions such as ``Are there any rubber things that have the same size as the yellow metallic cylinder?''.
That work is very powerful from the point of view of cognitive systems, vision and language.
Our approach is different because (i)~we focus on \emph{robotic} cognitive systems, including manipulation and the uncertainties inherent to robot vision and control, and (ii)~we follow the developmental paradigm and the embodiment hypothesis~\cite{lungarella:2003:devrobsurvey}, meaning that, leveraging the fact that a human and a humanoid produce actions with similar effects, we relate words with the robot's \emph{sensorimotor} experience, rather than sensory only~(purely images-to-text).

In robotics and cognitive systems research, both object-directed action recognition in external agents~\cite{koppula:2013:ijrr} and the incorporation of language in \hr{} systems~\cite{harnad:1990,matuszek:2014:aaai} have received ample attention, for example using the concept of \emph{intuitive physics}~\cite{lake:2017:bbs,gao:2018:acl} to be able to predict outcomes from real or simulated interactions with objects.
A growing interest is devoted to robots that learn new cognitive skills and improve their capabilities by interacting autonomously with the surrounding environment.
Robots operating in the real, unstructured world may understand available opportunities conditioned on their body, perception and sensorimotor experiences: the intersection of these elements gives rise to object \emph{affordances}~(action possibilities), as they are called in psychology~\cite{gibson:2014}.
The advantage of robot affordances lies in the ability to capture essential functional properties of environment objects in terms of the actions that the agent is able to perform with them, allowing to reason with prior knowledge about never-before-seen scenarios, thus exhibiting learning~\cite{montesano:2008,jamone:2016:tcds} and some degree of online adaptation~\cite{maestre:2017:icdl}.

Zech et al. published a systematic taxonomy of robot affordance models~\cite{zech:2017:ab}.
According to their criteria~(we refer the reader to the taxonomy for the precise definitions), in terms of \emph{perception} our work classifies as using
an agent perspective, meso-level features, $1$st~order, stable temporality;
in terms of \emph{development}: acquisition by exploration, prediction by inference, generalization
exploitation by action selection and language, offline learning.

Several works have studied the potential coupling between learning robot affordances and \emph{language grounding}.
The union of these two elements can give new skills to cognitive robots, such as:
creation of categorical concepts from multimodal association obtained by grasping and observing objects, while listening to partial verbal descriptions~\cite{nakamura:2009:iros,araki:2012:iros};
associating spoken words with sensorimotor experience~\cite{salvi:2012:smcb,morse:2016:cogsci};
linking language with sensorimotor representations~\cite{stramandinoli:2016:icdl}; or
carrying out complex tasks~(which require planning of a sequence of actions) expressed in natural language instructions to a robot~\cite{antunes:2016:icra}.

In particular Salvi et al.~\cite{salvi:2012:smcb}, which this paper extends, proposes a joint model to learn robot affordances~(i.e., relationships between actions, objects and resulting effects) together with word meanings.
The data used for learning such a model is from robot manipulation experiments, acquired from an ego-centric perspective.
Each experiment is associated with a number of alternative verbal descriptions uttered by two human speakers, for a total of \SI{1270}~recordings.
That framework assumes that the robot action is known \emph{a~priori} during the training phase~(e.g., during a grasping action the robot knows with certainty that it is performing a grasp), and the resulting model can be used at testing to make inferences about the environment.
In a recent work~\cite{saponaro:2017:glu} we relaxed the assumption of knowing the action.
We did this by merging the action estimation obtained from an external gesture recognizer~\cite{saponaro:2013:crhri} as \emph{hard evidence}~(i.e., certain evidence) to the full model, meaning that the action was deterministic.
By contrast, in this paper we propose a theoretical way to fuse the two sources of information~(about the self and about others) in a fully probabilistic manner, therefore introducing \emph{soft evidence}.
This addition allows to perform more fine-grained types of inferences and reasoning than before.
First, predictions over affordances and words when observing another agent with uncertainty.
Second, the generation of \emph{verbal descriptions} from the estimated word probabilities, for easier human interpretation of the model's explanations.

%%%%%%%%%%%%%%%%%%%%%%%%%%%%%%%%%%%%%%%%%%%%%%%%%%%%%%%%%%%%%%%%%%%%%%%%%%%%%%%%
%!TEX encoding = UTF-8 Unicode

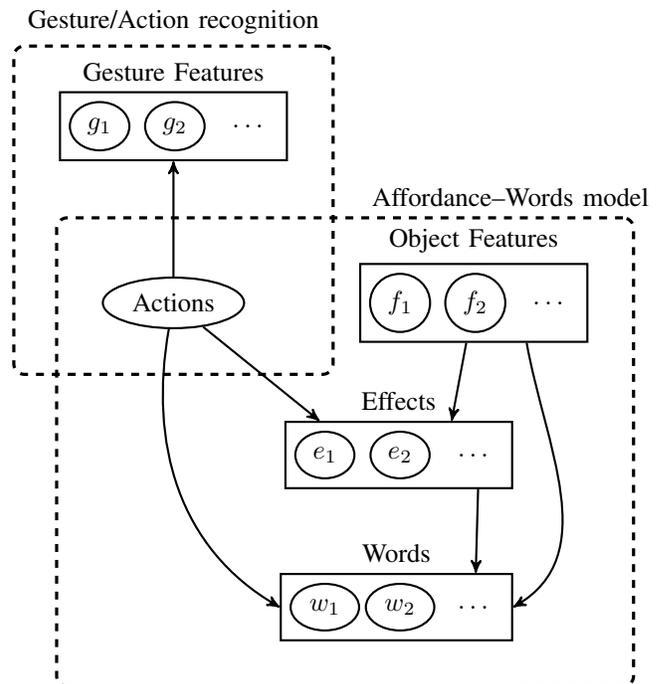
\begin{figure}
  \tikzstyle{affnode} = [ellipse, draw, thick]
  \tikzstyle{wordnode} = [ellipse, draw, thick]
  \tikzstyle{group} = [rectangle, draw, black, thick]
  \tikzstyle{dashedgroup} = [rectangle, draw, inner sep=0.6cm, dashed, rounded corners, black]
  \tikzstyle{dottedgrouptrapezium} = [trapezium, draw, dotted, rounded corners, black]
  \tikzstyle{affarrow} = [->, thick, >=stealth']
    \centering
    \begin{tikzpicture}
      % single nodes
      \node[affnode] (g1) {$g_1$};
      \node[affnode, right of=g1] (g2) {$g_2$};
      \node[right of=g2] (gdots) {$\dots$};
      \node[group, fit=(g1) (g2) (gdots),label=above:Gesture Features] (gestures) {};
      \node[affnode, below of=gestures] (actions) [below=1cm] {Actions};
      \node[affnode, right of=actions] (f1) [right=1.6cm] {$f_1$};
      \node[affnode, right of=f1] (f2) {$f_2$};
      \node[right of=f2] (fdots) {$\dots$};
      \node[affnode, below of=f1] (e2) [below=0.7cm] {$e_2$};
      \node[affnode, left of=e2] (e1) {$e_1$};
      \node[right of=e2] (edots) {$\dots$};
      \node[wordnode, below of=e1] (w1)  [below=0.7cm] {$w_1$};
      \node[wordnode, right of=w1] (w2) {$w_2$};
      \node[right of=w2] (wdots) {$\dots$};
      % groups
      \node[group, fit=(f1) (f2) (fdots),label=above:Object Features] (features) {};
      \node[group, fit=(e1) (e2) (edots),label=above:Effects] (effects) {};
      \node[group, fit=(w1) (w2) (wdots),label=above:Words] (words) {};
      \node[dashedgroup, fit=(actions) (features) (effects) (words),label={[shift={(0:2.2)}]above:\AffWords{} model}]{};
      % arrows
      \draw[affarrow] (actions) -- ([xshift=-30pt]effects.north);
      \draw[affarrow] (actions) to [out=260,in=150] (words.west);
      \draw[affarrow] (features) -- ([xshift=20pt]effects.north);
      \draw[affarrow] ([xshift=20pt]features.south) to [out=280,in=30] (words.east);
      \draw[affarrow] ([xshift=30pt]effects.south) -- ([xshift=30pt]words.north);
      % extra
      \draw[affarrow] (actions) -- (gestures);
      \node[dashedgroup, fit=(actions) (gestures),label=above:Gesture/Action recognition]{};
    \end{tikzpicture}
  \caption{Abstract representation of the probabilistic dependencies in the model.}
    \label{fig:model}
\end{figure}

\section{Method}
\label{sec:method}

The purpose of our work is to model the development of language learning from self-centered, individualistic learning to socially aware learning.
This transition happens gradually in subsequent phases.
In the first phase, the system engages in manipulation activities with objects in its environment~\cite{montesano:2008}.
The robot learns object affordances by associating object properties, actions and the corresponding effects.
In a second phase, the robot interacts with a human who uses spoken language to describe the robot's activities~\cite{salvi:2012:smcb}.
Here, the robot interprets the meaning of the words, grounding them in the \actionperception{} experience acquired so far.
Although this phase can already be considered \emph{social} for the presence of a human \emph{narrator}, it is still self-centered, because the robot is still learning how to interpret its own actions.
In the last phase, which is the contribution of this work, the system turns to observing human actions of a similar nature as the ones explored in the first phases (see examples in Fig.~\ref{fig:action_examples}).
The robot reuses the experience acquired in the first phases to interpret the new observations and to address the correspondence problem~\cite{nehaniv:2002:correspondence} between its own actions and the actions performed by the human.
In this phase, human movements are interpreted using the experience acquired so far, and they are incorporated into the model using a statistical gesture recognizer~\cite{saponaro:2013:crhri}.

Fig.~\ref{fig:model} illustrates the probabilistic dependencies in the complete model and will be detailed in the following subsections.

To permit the transfer from robot self-centered knowledge to human knowledge to work, we assume that the \emph{same actions}, performed on objects with the \emph{same properties}, cause the \emph{same effects} and are described by the \emph{same words}.
In other terms, all of the variables under consideration~(which will be described in Sec.~\ref{sec:experimental_settings}) are the link between robot and human.

In our theoretical formulation and in our implementation, we will hinge on the existence of the discrete Action variable, the value of which is known to the robot in the ego-centric phase of learning, but must be inferred when observing human actions.
This variable connects all the other observable variables in the model: human gesture features, object properties, effect variables and words.
This allows the robot to:
\begin{itemize}
\item use language in order to determine the mapping between human and own actions, and learn the corresponding perceptual models;

\item in many cases, use the affordance variables to infer the above mapping even in the absence of verbal descriptions;

\item once the perceptual models for human actions are acquired, use the complete model to do inference on any variable given some evidence.
\end{itemize}
In the remainder of this section, first we provide details, in Sec.~\ref{sec:bn}, about the probabilistic models enclosed in the \emph{\AffWords{} model} box of Fig.~\ref{fig:model}.
Then, in Sec.~\ref{sec:gesture_recognition} we describe the gesture recognition method.
Finally, in Sec.~\ref{sec:combination} we describe the way in which we combine evidence from the two models.

\subsection{\AffWords{} Model}
\label{sec:bn}
We use a Bayesian probabilistic framework to allow a robot to ground the basic world behavior and verbal descriptions associated to it.
All variables in the model are discrete or are discretized from continuous sensory variables through clustering in a preliminary learning phase.
The variables can be divided according to their use: action variable~$A = \{a\}$, object feature variables~$F=\{f_1, f_2, \dots\}$, effect variables~$E=\{e_1, e_2, \dots\}$ and word variables~$W = \{w_1, w_2, \dots\}$.
Details on the specific variables used in this study are given in Sec.~\ref{sec:experimental_settings}.

The \ac{BN} model~\cite{pearl:2014:probabilistic} relates all these variables by means of the joint probability distribution~$\pbn(A, F, E, W)$, sketched by the \AffWords{} model box in Fig.~\ref{fig:model}.
The dependency structure and the model parameters are estimated by the robot in an ego-centric way through interaction with the environment.
As a consequence, during learning, the robot knows what action it is performing with certainty, and the variable~$A$ assumes a deterministic value.
During inference, the probability distribution of the variable~$A$ can be inferred from evidence on the other variables.
For example, if the robot is asked to make a spherical object roll, it will be able to select the action tap as most likely to obtain the desired effect, based on previous experience.

\newcommand{\myscalefactor}{0.8}

\newcommand{\shapeOfHmmState}{circle}

\newcommand{\standardhmm}[1]{
    \node[draw,\shapeOfHmmState] (hmm#1s1) {$s_1$};
    \node[draw,\shapeOfHmmState, right of=hmm#1s1] (hmm#1s2) {$s_2$};
    \node[\shapeOfHmmState, right of=hmm#1s2] (hmm#1s3) {\dots};
    \node[draw,\shapeOfHmmState, right of=hmm#1s3] (hmm#1s4) {$s_Q$};
    \node[left of=hmm#1s1]  (invisible1) {};
    \node[right of=hmm#1s4] (invisible2) {};
    \path[->] (hmm#1s1) edge (hmm#1s2);
    \path[loop above] (hmm#1s1) edge (hmm#1s1);
    \path[->] (hmm#1s2) edge (hmm#1s3);
    \path[loop above] (hmm#1s2) edge (hmm#1s2);
    \path[dashed] (hmm#1s2) -- (hmm#1s3);
    \path[->] (hmm#1s3) edge (hmm#1s4);
    \path[loop above] (hmm#1s4) edge (hmm#1s4);
    \path[->] (invisible1) edge (hmm#1s1);
    \path[->] (hmm#1s4) edge (invisible2);
}

\newcommand{\modeltwo}{
  \begin{tikzpicture}[scale=\myscalefactor, every node/.style={scale=\myscalefactor}]
  \matrix (M) [matrix of nodes, ampersand replacement=\&] {%
    grasp gesture HMM \& \standardhmm{1} \\
    tap gesture HMM \& \standardhmm{2} \\
    touch gesture HMM \& \standardhmm{3} \\
  };
  \end{tikzpicture}
}

\subsection{Gesture Recognition}
\label{sec:gesture_recognition}
When observing a human performing an action, the value of the variable~$A$ is not known to the robot neither during learning nor during inference.
During learning, we assume that the robot has not yet acquired a perceptual model of the gestures associated to the human actions.
However, the value of~$A$ can be inferred, either from a verbal description of the scene, or from the other affordance variables through the \AffWords{} model described earlier.

For example, suppose that the \AffWords{} model predicts that performing a tap action on a spherical object will result in a high velocity of the object.
If the human performs an unknown action on a spherical object and obtains a high velocity, the robot will be able to infer that the action is most probably a tap, although it was not able to recognize the gesture associated with this action.

This information can be used to train our statistical \emph{gesture recognition system}~\cite{saponaro:2013:crhri}.
The system recognizes actions~(from gesture features) and corresponds to the Gesture/Action recognition block in Fig.~\ref{fig:model}.
It is based on \acp{HMM} with Gaussian mixture models as emission probability distributions.
Our input features are the~3D coordinates of the tracked human hand indicated by the~$g_i$ variables in Fig.~\ref{fig:model}.
The coordinates are transformed to be centered on the person torso~(to be invariant to the distance between the user and the sensor) and normalized to account for variability in amplitude~(to be invariant to wide/emphatic vs narrow/subtle executions of the same action).

The model for each action is a left-to-right \ac{HMM}, where the transition model between the~$Q$ discrete states~$\mathcal{S} = \{s_1, \dots, s_Q\}$ is structured so that states with a lower index represent events that occur earlier in time.

Although not expressed so far in the notation, the continuous variables~$g_i$ are measured at regular time intervals.
At a certain time step~$t$, the $D$-dimensional feature vector can be expressed as~$\bm{g}[t] = \{g_1[t], \dots, g_D[t]\}$.
The input to the model is a sequence of~$T$ such feature vectors~$\bm{g}[1], \dots, \bm{g}[T]$ that we call for simplicity~$G_1^T$, where~$T$ can vary for every recording.

At recognition~(testing) time, we can use the models to estimate the likelihood of a new sequence of observations~$G_1^T$ given each possible action, by means of the \FB{} inference algorithm.
We can express this likelihood as $\mathcal{L}_\text{HMM}(G_1^T \given A=a_k)$, where $a_k$ is one of the possible actions.
By normalizing the likelihoods, assuming that the gestures are equally likely \emph{a~priori}, we can obtain the posterior probability of the action given the sequence of observations as
\begin{equation} \label{eq:phmm_action}
  \phmm(A=a_k \given G_1^T) = \frac{\mathcal{L}_\text{HMM}(G_1^T \given A=a_k)}{\sum_h \mathcal{L}_\text{HMM}(G_1^T \given A=a_h)}.
\end{equation}

\subsection{Combining the \acs{BN} with Gesture \acsp{HMM}}
\label{sec:combination}
Once learned, the two models described above define two probability distributions over the relevant variables for the problem:
$\pbn(A, F, E, W)$ and~$\phmm(A \given G_1^T)$.
The goal during inference is to merge the information provided by both models and estimate~$\pcomb(A, F, E, W \given G_1^T)$, that is, the joint probability of all the affordance and word variables, given that we observe a certain action performed by the human.

To simplify the notation, we call $X = \{A, F, E, W\}$ the set of affordance and word variables~$\{a, f_1, f_2, \dots, e_1, e_2, \dots, w_1, w_2, \dots\}$.
During inference, we have a (possibly empty) set of observed variables~$\xobs \subseteq X$, and a set of variables $\xinf \subseteq X$ on which we wish to perform the inference.
In order for the inference to be non-trivial, it must be~$\xobs \cap \xinf = \varnothing$, that is, we should not observe any inference variable.
According to the \ac{BN} alone, the inference will compute the probability distribution of the inference variables~$\xinf$ given the observed variables~$\xobs$ by marginalizing over all the other (latent) variables $\xlat = X \setminus (\xobs \cup \xinf)$, where~$\setminus$~is the set difference operation:
\begin{equation*}
 \pbn(\xinf \given \xobs) = \sum_{\xlat} \pbn(\xinf, \xlat \given \xobs).
\end{equation*}

If we want to combine the evidence brought by the \ac{BN} with the evidence brought by the \ac{HMM}, there are two cases that can occur:
\begin{enumerate}
\item the action variable is included among the inference variables: $A \in \xinf$, or

\item the action variable is not included among the inference variables: $A \in \xlat$.
\end{enumerate}

Here, we are excluding the case where we observe the action directly~($A \in \xobs$) for two reasons.
First, this would correspond to the robot performing it by itself, whereas we are interested in interpreting other people's actions, which is a necessary skill to engage in social collaboration with humans.
Second, this would make the evidence on the gesture features~$G_1^T$ irrelevant, because in the model of Fig.~\ref{fig:model} there is a tail-to-tail connection~\cite{pearl:2014:probabilistic} from~$G_1^T$ to the rest of the variables through the action variable, which means that, given the action, all dependencies to the gesture features are dropped.

The two cases 1),~2) enumerated above can be addressed separately when we do inference.
In the first case, we call~$\xinf^\prime$ the set of inference variables excluding the action~$A$, that is, $\xinf = \{\xinf^\prime, A\}$.
We can write:
\begin{align} \label{eq:fusion_excluding_action}
  & \pcomb(\xinf \given  \xobs, G_1^T) = \pcomb(A, \xinf^\prime \given  \xobs, G_1^T)= \nonumber \\
  &= \sum_{\xlat} \pcomb(A, \xinf^\prime, \xlat \given \xobs, G_1^T)= \nonumber\\
  &= \sum_{\xlat} \left[\pbn(A, \xinf^\prime, \xlat \given \xobs, G_1^T)\right. \nonumber \\[-4mm]
    & \mspace{80mu} \left.\phmm(A, \xinf^\prime, \xlat \given \xobs, G_1^T)\right]= \nonumber \\
  &= \left[\sum_{\xlat} \pbn(A, \xinf^\prime, \xlat \given \xobs)\right] \phmm(A \given G_1^T)= \nonumber \\
  &= \pbn(\xinf \given \xobs) \phmm(A \given G_1^T).
\end{align}
This means that we can evaluate the two models independently, then multiply the distribution that we obtain from the \ac{BN}~(over all the possible value of the inference variables) by the \ac{HMM} posterior for the corresponding value of the action.

In the second case, where the action is among the latent variables, we define, similarly, $\xlat = \{A, \xlat^\prime\}$, and we have:
\begin{align} \label{eq:fusion_including_action}
  & \pcomb(\xinf \given \xobs, G_1^T) = \nonumber \\
  &= \sum_{\{A,\xlat^\prime\}} \pcomb(\xinf, A, \xlat^\prime \given \xobs, G_1^T)= \nonumber \\
  &= \sum_{\{A,\xlat^\prime\}} \left[\pbn(\xinf, A, \xlat^\prime \given \xobs, G_1^T)\right. \nonumber \\[-4mm]
    & \mspace{100mu} \left.\phmm(\xinf, A, \xlat^\prime \given \xobs, G_1^T)\right]= \nonumber \\[2mm]
  &= \sum_{\{A,\xlat^\prime\}} \left[\pbn(\xinf, A, \xlat^\prime \given \xobs) \phmm(A \given G_1^T)\right]= \nonumber \\
  &= \sum_{A}\left[\phmm(A \given G_1^T)\sum_{\xlat^\prime} \pbn(\xinf, A, \xlat^\prime \given \xobs)\right]= \nonumber \\
  &= \sum_{A}\left[\phmm(A \given G_1^T) \pbn(\xinf, A \given \xobs)\right].
\end{align}
This time, we first need to use the \ac{BN} to do inference on the variables~$\xinf$ and~$A$, and then we marginalize out the action variable~$A$ after having multiplied the probabilities by the \ac{HMM} posterior.

\subsection{Generation and Scoring of Verbal Descriptions}
\label{sec:method:verbal}

In order to illustrate the language capabilities of the model, rather than displaying the probability distribution of the words inferred by the model, we use the \ac{CFG} described in Appendix~\ref{appendix:grammar} to generate written descriptions of the robot observations, on the basis of those probabilities.
Note that this grammar is defined here with the only purpose of interpreting the probability distributions over the words.
In the \AffWords{} model that we use, the speech recognizer is based on a free loop of words with uniform prior, and the Bayesian model relies on a bag-of-words assumption.
No grammatical (syntactic) information about the spoken descriptions was, therefore, used during learning.

In the current study, by merging the \AffWords{} model and the gesture recognition model, we allow the robot to \emph{reinterpret} the concepts it has learned in the self-centered phase, but we do not add any new words to the model.
Consequently, the descriptions that the model generates when observing humans use the same words to describe the agent (see also Sec.~\ref{sec:results:verbal_descriptions}).

The textual descriptions are generated as follows: given some evidence~$\xobs$ that we provide to the model and some human observation features~$G_1^t$ extracted from frames~$1$ to~$t$, we extract the generated word probabilities
$P(w_i \given \xobs, G_1^t)$.
We generate~$N$ sentences randomly from the \ac{CFG} using the \texttt{HSGen} tool from HTK~\cite{young:htkbook}.
Then, the sentences are re-scored according to the log-likelihood of each word in the sentence, normalized by the length of the sentence:
\begin{equation} \label{eq:sentence_score}
  \text{score}(s_j \given \xobs, G_1^t) = \frac{1}{L_j} \sum_{k=1}^{L_j} \log P(w_{jk} \given \xobs, G_1^t),
\end{equation}
where~$s_j$ is the~$j$th sentence,~$L_j$ is the number of words in the sentence~$s_j$, and~$w_{jk}$ is the~$k$th word in the sentence~$s_j$.
Finally, an $N$-best list of possible descriptions is produced by sorting the scores.

%%%%%%%%%%%%%%%%%%%%%%%%%%%%%%%%%%%%%%%%%%%%%%%%%%%%%%%%%%%%%%%%%%%%%%%%%%%%%%%%
%!TEX encoding = UTF-8 Unicode

\section{Experimental Settings}
\label{sec:experimental_settings}
Our experiments consist on testing our method on a number of example scenarios that will be described in Sec.~\ref{sec:results}.
In this section we provide experimental details and key assumptions of the method.

\begin{table}
    \centering
    \caption{The symbolic variables of the \acl{BN} (from~\cite{salvi:2012:smcb}), with the corresponding discrete values obtained from clustering during robot exploration of the environment.
    We call \emph{word variables} the booleans of the last row, whereas we call \emph{affordance variables} all the other symbols.}
    \label{tab:bnsymb}
    \begin{tabular}{cp{3.3cm}l}
    \toprule
    symbol & name: description     & values \\
    \midrule
    $a$    & Action: motor action  & grasp, tap, touch \\
    \midrule
    $f_1$  & Color: object color   & blue, yellow, green1, green2 \\
    $f_2$  & Size: object size     & small, medium, big \\
    $f_3$  & Shape: object shape   & sphere, box \\ % in the implementation the value is called circle, but sphere is clearer
    \midrule
    $e_1$  & ObjVel: object velocity                     & slow, medium, fast \\
    $e_2$  & HandVel: robot hand velocity                & slow, fast \\
    $e_3$  & ObjHandVel: relative \objecthand{} velocity & slow, medium, fast \\
    $e_4$  & Contact: object hand contact                & short, long \\
    \midrule
    $w_1$--$w_{49}$ & presence of each word in the verbal description & true, false \\
    \bottomrule
    \end{tabular}
\end{table}

\subsection{\AffWords{} Model}
Table~\ref{tab:bnsymb} presents a list of variables and the corresponding values used in the \AffWords{} model.
Note that the name of the values of the affordance variables have been assigned by us arbitrarily to the clusters, for the sake of making the results more human-interpretable.
However, the robot has no prior knowledge about the meaning of these clusters nor about their order, in case they correspond to ordered quantities.
For extracting object features and effects from the sensory data, we assume that the robot possesses visual segmentation and geometric reasoning capabilities, meaning that it is able to segment the~(potentially multiple) regions of interest corresponding to the physical objects of the world from the background~(e.g., a planar surface such as a table) and to determine their positions.

We use the following notation in order to distinguish between the values of the affordance variables~(all but the last row in Table~\ref{tab:bnsymb}) and the words~(last row in the table).
Words and sentences are always enclosed in quotation marks.
For example, ``sphere'' will refer to the spoken word, whereas sphere will refer to the value of the Shape variable corresponding to the specific cluster.
Similarly, ``grasp'' will correspond to a spoken word, whereas grasp corresponds to a value of the Action variable.

There is no one-to-one correspondence between the values of the affordance variables and words.
This was partly emerging from the natural variability that is inherent in the way humans describe situations in spoken words.
It was also a design choice, because we wanted to prove that the model was not merely able to recover simple \wordmeaning{} associations, but was able to cope with more natural spoken utterances.
Consequently, in the spoken descriptions:
(i)~there are many synonyms for the same concept: for instance, cubic objects are called ``box'', ``square'' or ``cube''. Also, actions and effects are described using different tenses (``is grasping'', ``grasped'', ``has (just) grasped'');
(ii)~different affordance variable values may have the same associated verbal description, e.g., two color clusters corresponding to different shades of green are both referred to as ``green'';
(iii)~finally, many affordance variable values have no direct description: for example, the object velocity and \objecthand{} velocity~(slow, medium, fast), or the \objecthand{} contact~(short, long) are never described directly, and need to be inferred from the situation.

The \AffWords{} model does not account for the concepts of parts of speech, verb tenses or \emph{temporal aspects} explicitly.
For example, the words ``is'', ``grasping'', ``has'', ``grasped'', ``just'', and so on, are initially completely distinct and unrelated to the model, which has no prior information about what verbs, adjectives or nouns are, nor about similarity between words.
It is only through the association with the other robot observations that the model realizes that ``grasping'' has the same meaning as ``grasped''.
The following three phrases, which were used interchangeably in the experiments, are mapped to exactly the same meaning, after learning:
(i)~``is grasping'',
(ii)~``has grasped'',
(iii)~``grasped''.
Note that the model \emph{per~se} would be fully capable to distinguish between those phrases, provided that they were used in different situations, which however was not the case in our experimental data.

\subsection{Gesture Recognition}
\label{sec:experimental_settings:gesture_recognition}
In this work, we consider three independent, multiple-state \aclp{HMM}, each of them trained to recognize one of the considered manipulation gestures of  Fig.~\ref{fig:action_examples}.
The 3D coordinates of the human limbs and torso used to extract the input to the gesture recognizer are obtained with a commodity depth sensor~(Kinect)\footnote{Currently, our gesture recognition algorithm relies on human skeleton tracking software from a depth stream.
In our experience, the hand tracking is not reliable in the presence of a tabletop~(i.e., partially occluded human) as in Fig.~\ref{fig:action_examples}, so we record the same gestures twice, with and without the table: the latter is used for ensuring the robustness of the estimated hand coordinate, the former is used throughout the rest of our model and experiments.
We plan to overcome this limitation in future work.}.

%%%%%%%%%%%%%%%%%%%%%%%%%%%%%%%%%%%%%%%%%%%%%%%%%%%%%%%%%%%%%%%%%%%%%%%%%%%%%%%%
%!TEX encoding = UTF-8 Unicode

\begin{figure}
\centering
\includegraphics[width=0.9\columnwidth]{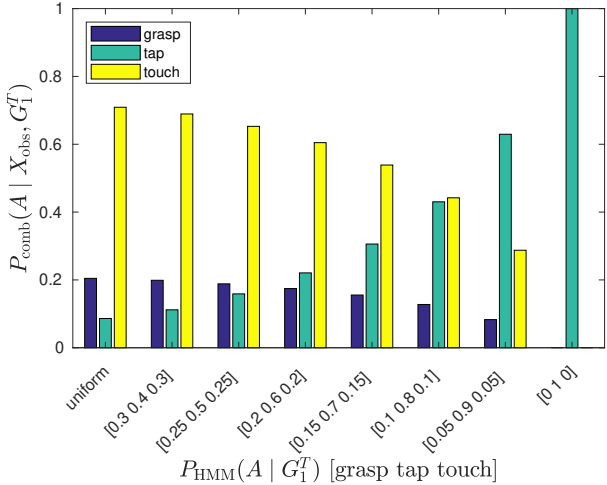}
\caption{Inference over action given the evidence $\xobs =\{\text{Size}=\text{small}, \text{Shape}=\text{sphere}, \text{ObjVel}=\text{slow}\}$, combined with different probabilistic soft evidence about the action.}
\label{fig:impact_of_evidence_on_Action}
\end{figure}

\section{Results}
\label{sec:results}

In this section, we report the experimental findings obtained with our proposed model.
Because it is based on \aclp{BN}, the model can make inferences over any set of its variables~$\xinf$, given any other set of observed variables~$\xobs$.
In particular, the model can do reasoning on the elements that constitute our computational concept of affordances, i.e., Action, Object Features, Effects in Fig.~\ref{fig:model}.
Furthermore, it can do reasoning over Words.
We present the following types of results:
\begin{itemize}
  \item inferences over affordance variables~(see Table~\ref{tab:bnsymb}) in Sec.~\ref{sec:results:inference_actions}, \ref{sec:results:inference_effects}, \ref{sec:results:anticipation_effects};

  \item predictions of word probabilities in Sec.~\ref{sec:results:prediction_words};

  \item verbal descriptions generated from the word probabilities of the previous point, according to a formal grammar. The descriptions, in turn, can be interpreted to observe the emergence of certain language phenomena: Sec.~\ref{sec:results:verbal_descriptions}, \ref{sec:results:conjunction}, \ref{sec:results:description_objects}.
\end{itemize}

\begin{figure*}
\centering
\subfloat[][Predictions with a sphere object.]
{ \includegraphics[width=0.45\linewidth]{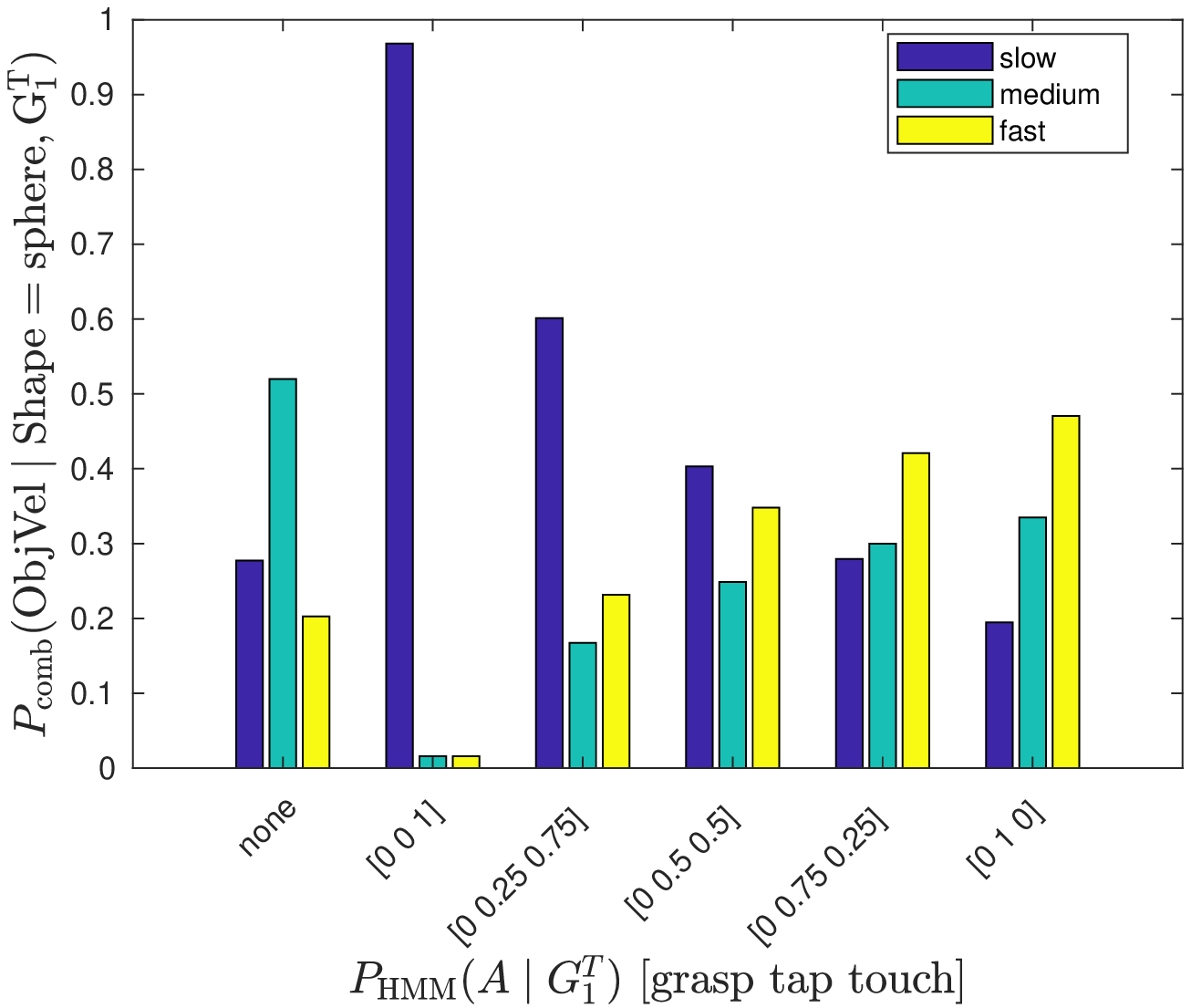} \label{fig:impact_of_evidence_on_ObjVel_sphere} } \quad
\subfloat[][Predictions with a box object.]
{ \includegraphics[width=0.45\linewidth]{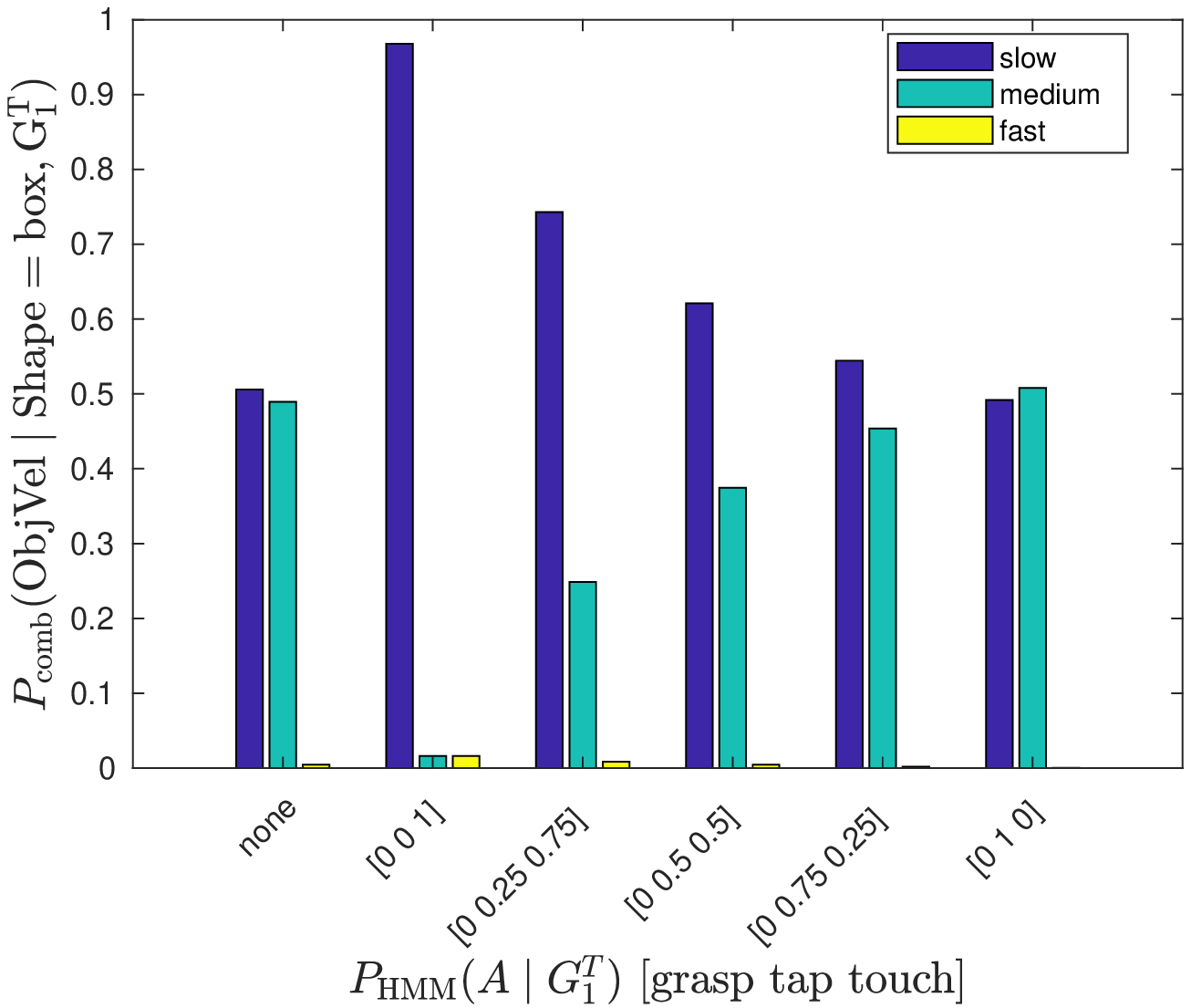} \label{fig:impact_of_evidence_on_ObjVel_box} }
\caption{Inference over the object velocity effect of different objects, when given probabilistic soft evidence about the action.}
\label{fig:impact_of_evidence_on_ObjVel}
\end{figure*}

\begin{figure*}
  \centering
  \subfloat[][Action performed on small sphere. Description: ``the robot pushed the ball and the ball moves''.]{
    \resizebox{0.9\linewidth}{!}{
      \begin{tikzpicture}
        \node (lik) {\includegraphics[width=0.6\linewidth]{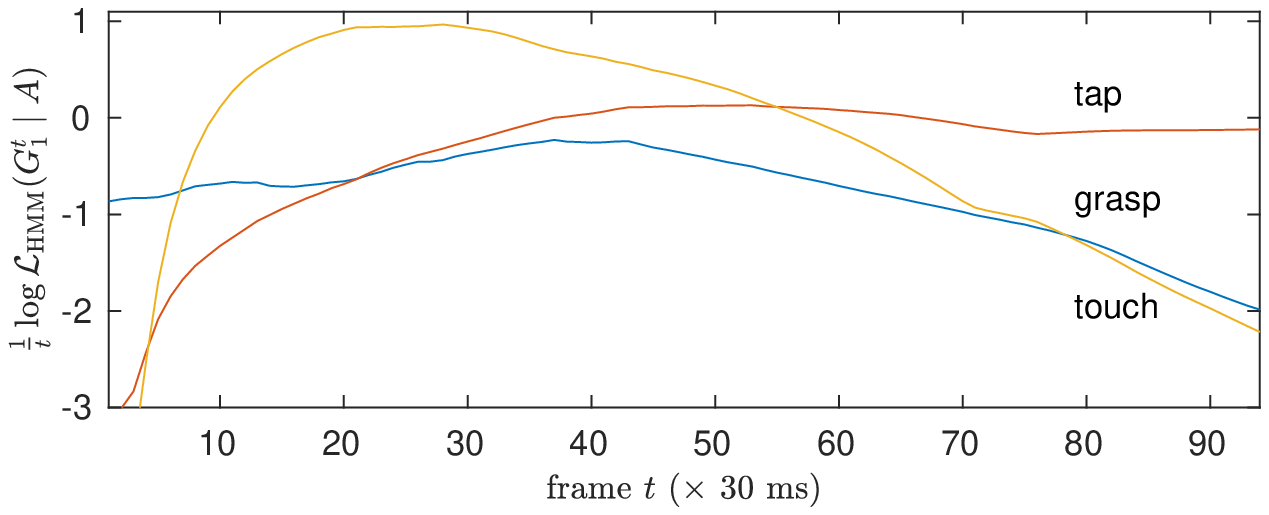}};
        \node at ([xshift=-90pt,yshift=30pt]lik.north) {\includegraphics[width=\myWidth\linewidth]{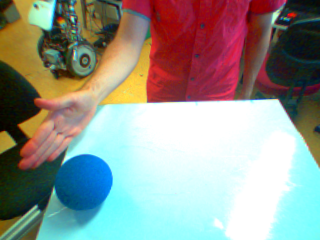}};
        \node at ([xshift=+10pt,yshift=30pt]lik.north) {\includegraphics[width=\myWidth\linewidth]{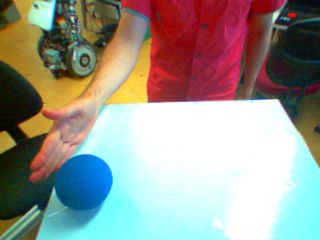}};
        \node at ([xshift=+110pt,yshift=30pt]lik.north) {\includegraphics[width=\myWidth\linewidth]{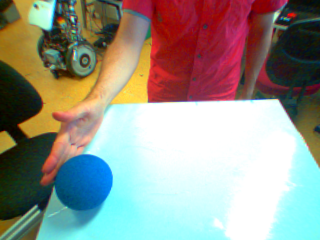}};
        \node at ([xshift=100pt,yshift=30pt]lik.east) {\includegraphics[width=0.35\linewidth]{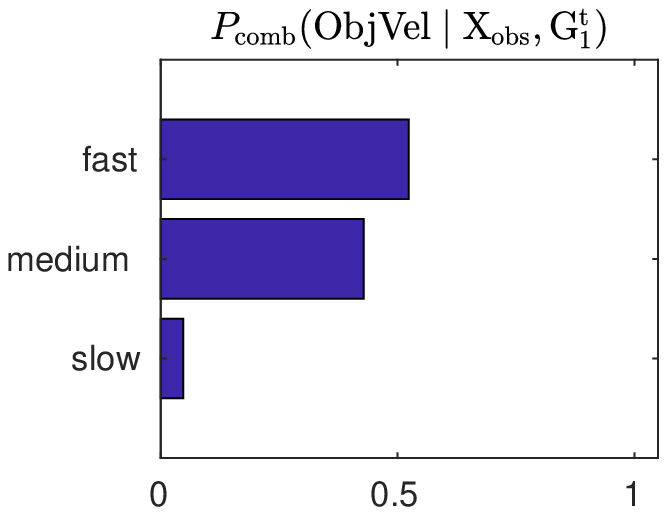}};
      \end{tikzpicture}
    } % end resizebox
    \label{fig:effect_pred_sphere}
  } % end subfloat

  \subfloat[][Action performed on big box. Description: ``the robot is pushing the big square but the box is inert''.]{
    \resizebox{0.9\linewidth}{!}{
      \begin{tikzpicture}
        \node (lik) {\includegraphics[width=0.6\linewidth]{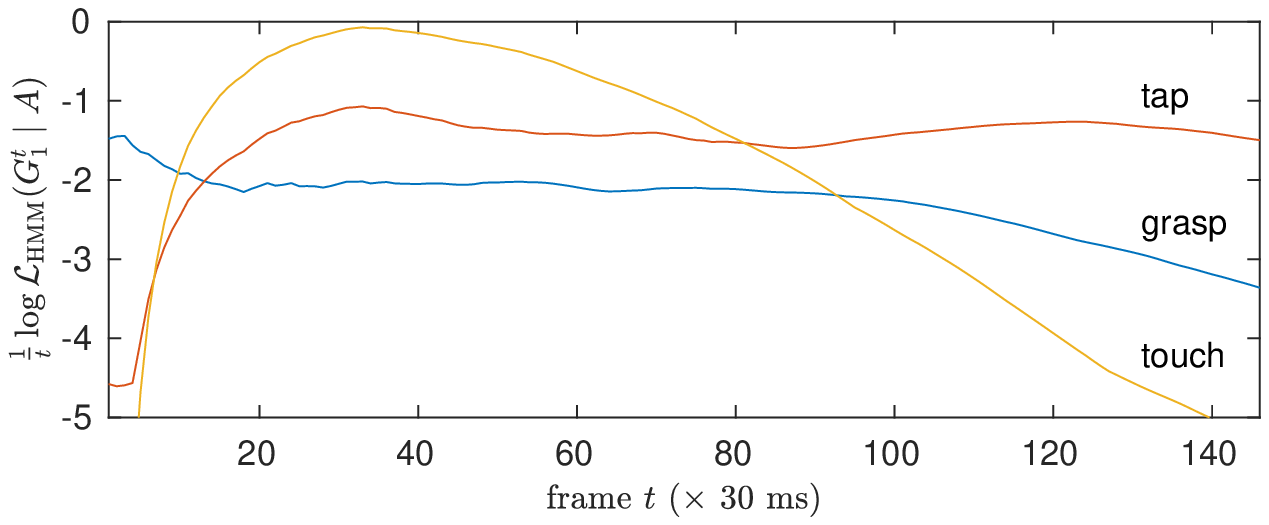}};
        \node at ([xshift=-90pt,yshift=30pt]lik.north) {\includegraphics[width=\myWidth\linewidth]{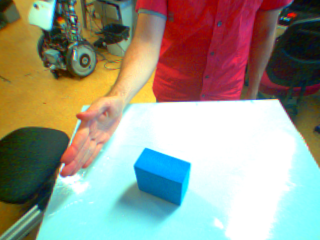}};
        \node at ([xshift=+10pt,yshift=30pt]lik.north) {\includegraphics[width=\myWidth\linewidth]{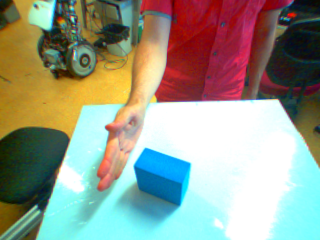}};
        \node at ([xshift=+110pt,yshift=30pt]lik.north) {\includegraphics[width=\myWidth\linewidth]{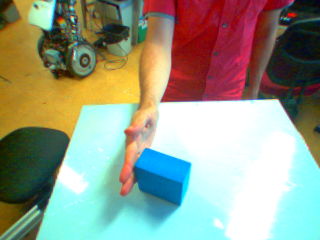}};
        \node at ([xshift=100pt,yshift=30pt]lik.east) {\includegraphics[width=0.35\linewidth]{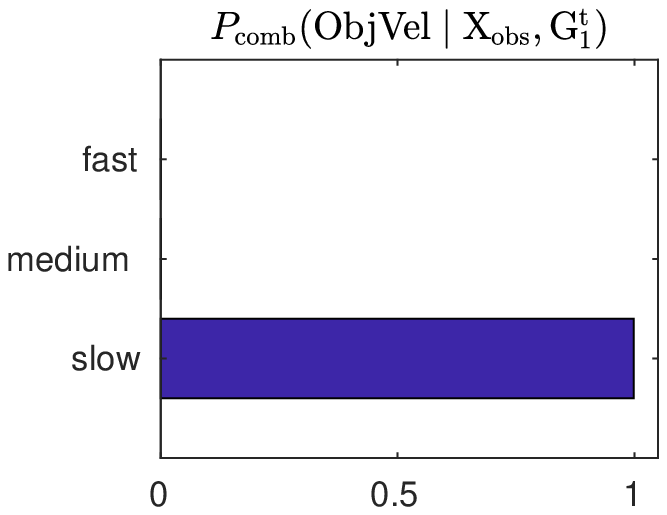}};
      \end{tikzpicture}
    } % end resizebox
  \label{fig:effect_pred_box}
  } % end subfloat
  \caption{Object velocity effect anticipation before impact. The evidence from the gesture recognizer~(left) is fed into the \AffWords{} model before the end of the execution. The combined model predicts the effect~(right) and describes it in words.}
  \label{fig:effect_pred}
\end{figure*}

\subsection{Action Recognition}
\label{sec:results:inference_actions}

In this experiment, we test the ability of our approach to recognize actions.
Both the \AffWords{} model and the gesture recognition model can each perform inference of the Action variable individually: the former by using the variables of Tab.~\ref{tab:bnsymb}, the latter by using human gesture features.
We show how our method performs the inference over Action in a joint way.
This includes dealing with information with different degrees of confidence, or conflicting information.

The scene of Fig.~\ref{fig:impact_of_evidence_on_Action} contains a small ball which, after the manipulative action, exhibits a low velocity.
Based on the evidence, the affordance model gives the highest probability $\pbn(A \given \xobs)$ to the action \emph{touch}, which usually does not result in any movement of the object.
However, in this particular simulated situation, we assume that the action performed by the human was an (unsuccessful) \emph{tap}, that is, a tap that does not result in any movement for the object.
In the simulation we show the effect of augmenting the inference with information from a gesture recognizer, that is, computing $\pcomb(A \given \xobs,G_1^T)$.
We analyze the effect of varying the degree of confidence of the classifier.
We start from a uniform posterior $\phmm(A \given G_1^T)$, corresponding to a poor classifier, and gradually increase the probability of the correct action until it reaches~$1$.
In this particular example, in order to win the belief of the affordance model, the action recognition needs to be very confident ($\phmm(A=\text{tap} \given G_1^T) > 0.81$).

\begin{figure}
\centering
\includegraphics[width=0.9\columnwidth]{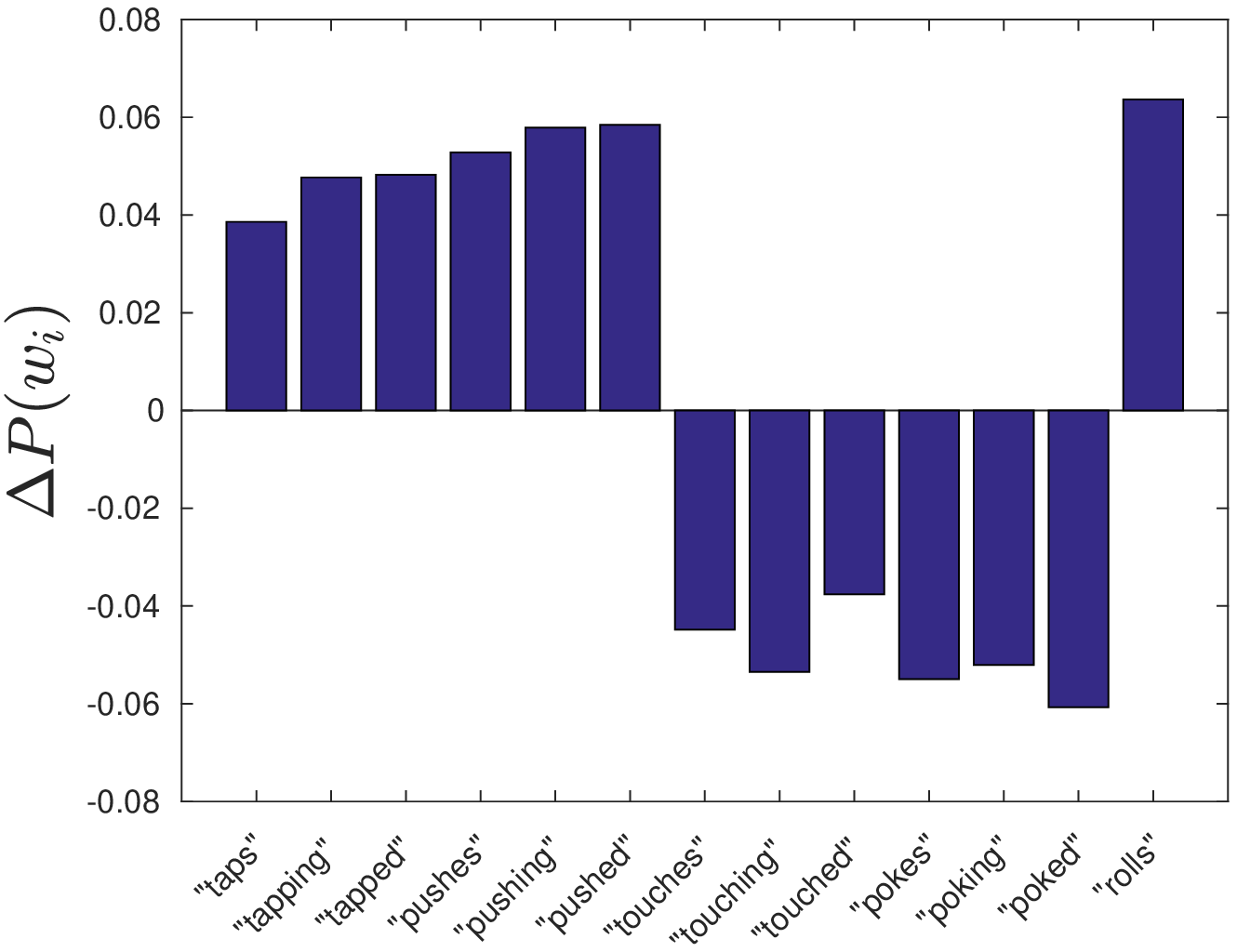}
\caption{Variation of word occurrence probabilities:
$\Delta P(w_i) = \pcomb(w_i \given \xobs, \text{Action=tap}) - \pbn(w_i \given \xobs)$, where $\xobs = \{ \text{Size=big, Shape=sphere, ObjVel=fast} \}$.
This variation corresponds to the difference of word probability when we add the tap action evidence~(obtained from gesture recognition) to the initial evidence about object features and effects. We have omitted words for which no significant variation was observed.}
\label{fig:probdiff}
\end{figure}

\subsection{Effect Prediction}
\label{sec:results:inference_effects}

We now show how our approach does inference over a different variable~(instead of the Action one which is common between \AffWords{} model and gesture model), i.e., how it predicts the value of the object velocity effect variable.
We will do this by using different degrees of probabilistic confidence about the action, and analyzing the outcome in terms of velocity prediction.
This experiment exposes that \emph{all} the variables of Tab.~\ref{tab:bnsymb} jointly link robot and human, not only the Action variable, for the reasons expressed in Sec.~\ref{sec:method}.

Fig.~\ref{fig:impact_of_evidence_on_ObjVel} shows the considered inference in two cases: when the prior information says that the shape is spherical~(see Fig.~\ref{fig:impact_of_evidence_on_ObjVel_sphere}), and when it is cubic~(see Fig.~\ref{fig:impact_of_evidence_on_ObjVel_box}).

The leftmost distribution in both figures shows the prediction of object velocity from the \AffWords{} model alone, without any additional information.
When the shape is spherical, the model is not sure about the velocity, whereas if the shape is cubic, the model does not expect high velocities.
If we add clear evidence on the action \emph{touch} from the action recognition model, suddenly the combined model predicts slow velocities in both cases, as expected.
However, if the action recognition evidence is gradually changed from \emph{touch} to \emph{tap}, the predictions of the model depend on the shape of the object.
Higher velocities are expected for spherical objects that can roll, compared to cubic objects.

\subsection{Effect Anticipation}
\label{sec:results:anticipation_effects}

Since the gesture recognition method interprets sequences of human motions, we can test this predictive ability of the complete model when we observe an incomplete action.
Fig.~\ref{fig:effect_pred} shows an example of this where we reason about the expected object velocity caused by a tap action.
Fig.~\ref{fig:effect_pred_sphere} shows the action performed on a spherical object, whereas Fig.~\ref{fig:effect_pred_box} on a cubic one.
The graphs on the left side show the time evolution of the evidence $\phmm(A \given G_1^t)$ from the gesture recognition model.
In order to make the variations emerge more clearly, instead of the posterior, we show $\frac{1}{t} \log \mathcal{L}_\text{HMM} (G_1^t \given A)$: the log-likelihood normalized by the length of the sequence.
Note how, in both cases, the correct action is recognized by the model given enough evidence, although the observation sequence is not complete.
The right side of the plot shows the prediction of the object velocity, given the incomplete observation of the action and the object properties.
The model correctly predicts that the sphere will probably move but the box is unlikely do so.
Finally, the captions in the figure also show the verbal description~(see Sec.~\ref{sec:method:verbal}) generated by feeding the probability distribution of the words estimated by the model given the evidence into the \acl{CFG}.

\begin{table}
    \centering
    \caption{$10$-best list of sentences generated from the evidence $\xobs = \{ \text{Color=yellow, Size=big, Shape=sphere, ObjVel=fast} \}$.}
    \label{tab:example_generated_sentences}
    \resizebox{\linewidth}{!}{% https://tex.stackexchange.com/a/27105
    \begin{tabular}{ll}
    \toprule
    sentence & score \\
    \midrule
    ``the robot pushed the ball and the ball moves'' & $-0.54322$ \\ % not good to have push
    ``the robot tapped the sphere and the sphere moves'' & $-0.5605$ \\
    ``he is pushing the sphere and the sphere moves'' & $-0.57731$ \\
    ``the robot is tapping the yellow ball and the big yellow sphere is moving'' & $-0.57932$ \\
    ``he pushed the yellow ball and the sphere is rolling'' & $-0.58853$ \\
    ``the robot is poking the ball and the sphere is rolling'' & $-0.58998$ \\
    ``he is pushing the ball and the yellow ball moves'' & $-0.59728$ \\
    ``he pushes the sphere and the ball is moving'' & $-0.60528$ \\
    ``he is tapping the yellow ball and the ball is moving'' & $-0.60675$ \\
    ``the robot pokes the sphere and the ball is rolling'' & $-0.60694$ \\
    \bottomrule
    \end{tabular}%
    } % end resizebox
\end{table}

\subsection{Prediction of Word Probabilities}
\label{sec:results:prediction_words}

Our model permits to make predictions over the word variables associated to affordance evidence.
In Fig.~\ref{fig:probdiff} we show the variation in word occurrence probabilities between two cases:
\begin{enumerate}
\item when the robot's prior knowledge evidence consists of information about object features and effects only: \{Size=big, Shape=sphere, ObjVel=fast\};

\item when the evidence corresponds to the one of the previous point, with the addition of the \emph{tap} action observed from the gesture recognizer (hard evidence).
\end{enumerate}

In this result, we notice two facts.
First, the probabilities of words related to tapping and pushing increase when a tapping action evidence from gesture recognition is introduced; conversely, the probabilities of other action words~(touching and poking) decreases.
Second, the probability of the word ``rolling''~(which is an effect of an action onto an object) also increases when the tap action evidence is entered.

\newcommand{\evidenceProducingAnd}{$\xobs=$\{ Action=grasp, ObjVel=medium \}}
\newcommand{\evidenceProducingBut}{$\xobs=$\{ Action=grasp, ObjVel=slow \}}

\begin{figure*}
  \centering
  \subfloat[][Evidence: \evidenceProducingAnd.]{
    \begin{tabular}[b]{c}
    \includegraphics[width=0.4\linewidth]{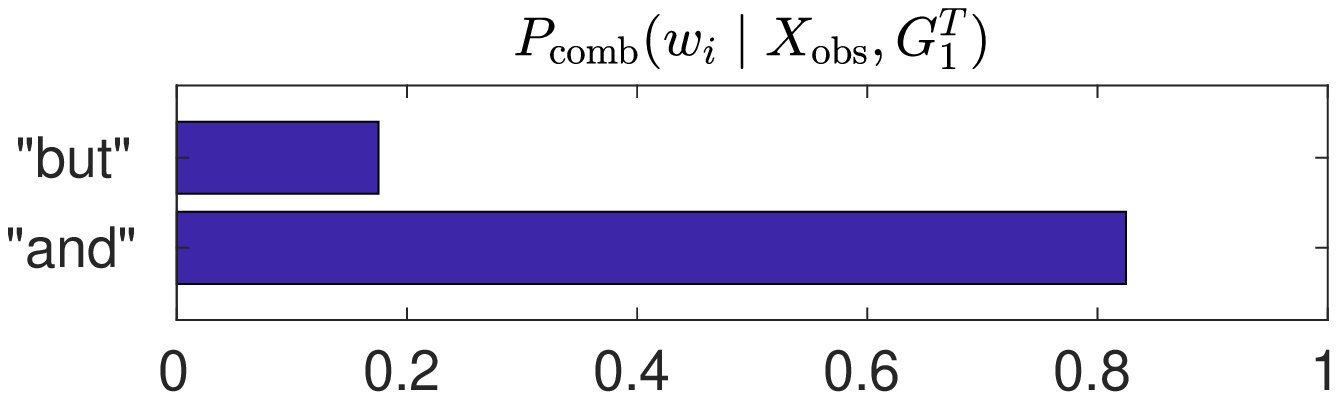}\\

    \resizebox{!}{0.1\linewidth}{% https://tex.stackexchange.com/a/27105
      \begin{tabular}{ll}
        \toprule
        sentence & score \\
        \midrule
        ``the robot is picking the sphere \textbf{and} the sphere is moving''  & $-0.59328$ \\
        ``the robot grasps the sphere \textbf{and} the ball is moving''  & $-0.59507$ \\
        ``the robot is picking the sphere \textbf{and} the sphere is rising''  & $-0.60882$ \\
        ``the robot grasped the sphere \textbf{and} the sphere is rising''  & $-0.61842$ \\
        ``the robot picked the ball \textbf{and} the ball is rising''  & $-0.64052$ \\
        ``baltazar grasps the sphere \textbf{and} the sphere is moving''  & $-0.66182$ \\
        ``the robot has grasped the ball \textbf{and} the ball is rising''  & $-0.66398$ \\
        ``the robot picked the ball \textbf{and} the green ball is moving''  & $-0.67134$ \\
        ``baltazar grasped the sphere \textbf{and} the ball is moving''  & $-0.67283$ \\
        ``baltazar is grasping the ball \textbf{and} the sphere is rising''  & $-0.6787$ \\
        \bottomrule
      \end{tabular}%
    } % end resizebox
    \end{tabular}
    \label{tab:conjunction:and}
  } % end subfloat
  \subfloat[][Evidence: \evidenceProducingBut.]{
    \begin{tabular}[b]{c}
    \includegraphics[width=0.4\linewidth]{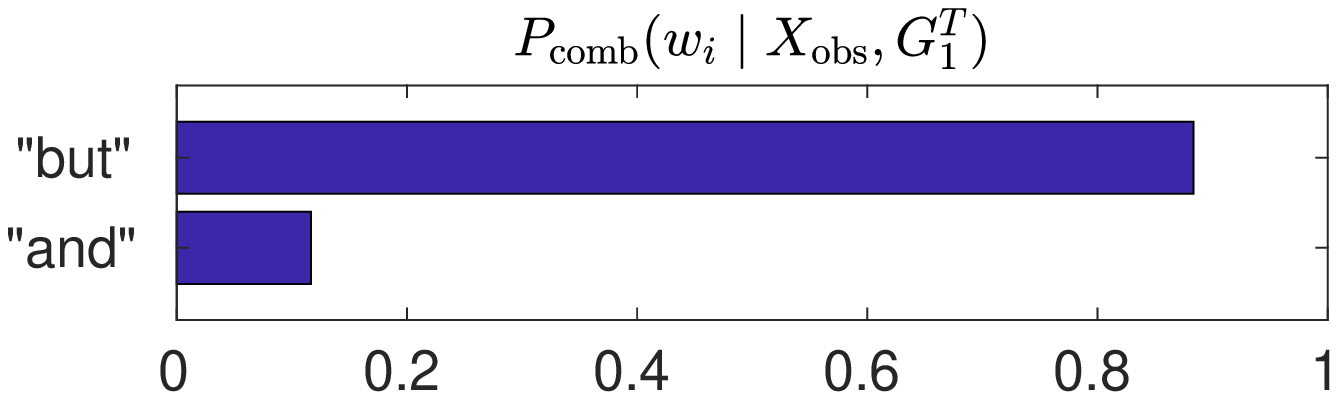}\\

    \resizebox{!}{0.1\linewidth}{% https://tex.stackexchange.com/a/27105
      \begin{tabular}{ll}
        \toprule
        sentence & score \\
        \midrule
        ``the robot is picking the cube \textbf{but} the square is still''  & $-0.52575$ \\
        ``the robot is grasping the sphere \textbf{but} the box is inert''  & $-0.55$ \\
        ``the robot is grasping the square \textbf{but} the sphere is still''  & $-0.55388$ \\
        ``the robot grasped the square \textbf{but} the cube is inert''  & $-0.55608$ \\
        ``baltazar is grasping the square \textbf{but} the square is inert''  & $-0.5571$ \\
        ``the robot is grasping the cube \textbf{but} the ball is inert''  & $-0.56011$ \\
        ``the robot picks the box \textbf{but} the square is inert''  & $-0.56397$ \\
        ``baltazar is picking the square \textbf{but} the square is still''  & $-0.56402$ \\
        ``he is grasping the square \textbf{but} the cube is inert''  & $-0.56815$ \\
        ``the robot grasps the square \textbf{but} the sphere is inert''  & $-0.57417$ \\
        \bottomrule
      \end{tabular}%
    } % end resizebox
    \end{tabular}
    \label{tab:conjunction:but}
  } % end subfloat
    \caption{$10$-best list of sentences generated given two different sets of evidence.
    In~(a) the model interprets the object movement as indicating a succesful grasp and uses the conjunction ``and''.
    In~(b) the slow movement is interpreted as no movement at all and, therefore, as an unsuccessful grasp: for that reason, the conjunction ``but'' is used.}
    \label{tab:conjunction}
\end{figure*}

\newcommand{\graspBoxGreenOne}{``the robot is grasping the box and the green box is moving''}
\newcommand{\touchBoxGreenOne}{``the robot is poking the green square and the cube is inert''}
\newcommand{\graspSphereGreenTwo}{``the robot picked the ball and the green ball is moving''}
\newcommand{\touchSphereGreenTwo}{``baltazar is poking the green sphere and the sphere is still''}

\newcommand{\evidenceProducingGraspBoxGreenOne}{$\xobs = \{ \text{Action=grasp, Color=green1, Shape=box} \}$}
\newcommand{\evidenceProducingTouchBoxGreenOne}{$\xobs = \{ \text{Action=touch, Color=green1, Shape=box} \}$}
\newcommand{\evidenceProducingGraspSphereGreenTwo}{$\xobs = \{ \text{Action=grasp, Color=green2, Shape=sphere} \}$}
\newcommand{\evidenceProducingTouchSphereGreenTwo}{$\xobs = \{ \text{Action=touch, Color=green2, Shape=sphere} \}$}

\begin{figure}
  \centering
  \subfloat[][\graspBoxGreenOne.]{
    \resizebox{\linewidth}{!}{
      \includegraphics{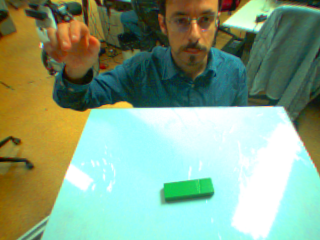}
      \includegraphics{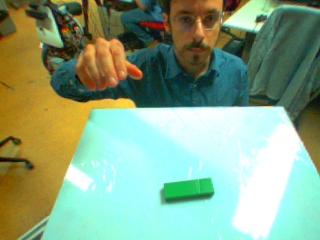}
      \includegraphics{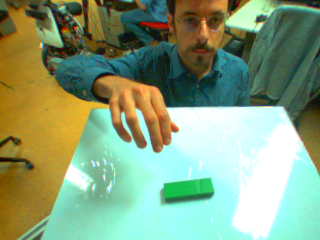}
      \includegraphics{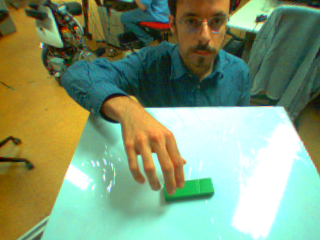}
      \includegraphics{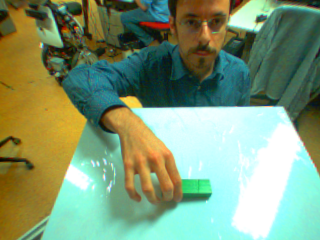}
      \includegraphics{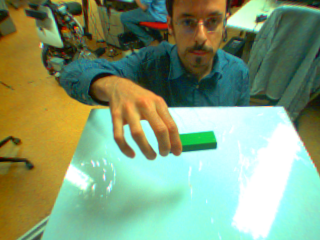}
    } % end resizebox
    \label{fig:descriptions:graspBoxGreenOne}
  } % end subfloat

  \subfloat[][\touchBoxGreenOne.]{
    \resizebox{\linewidth}{!}{
      \includegraphics{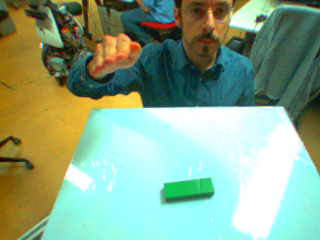}
      \includegraphics{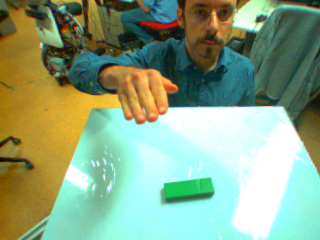}
      \includegraphics{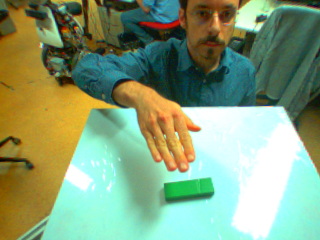}
      \includegraphics{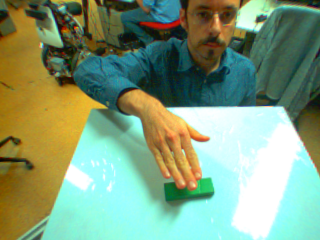}
      \includegraphics{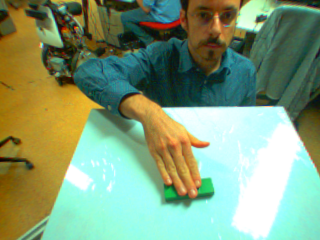}
      \includegraphics{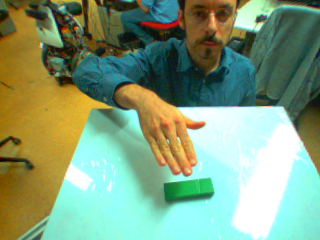}
    } % end resizebox
    \label{fig:descriptions:touchBoxGreenOne}
  } % end subfloat

  \subfloat[][\graspSphereGreenTwo.]{
    \resizebox{\linewidth}{!}{
      \includegraphics{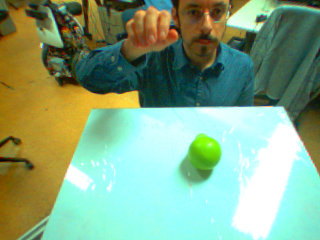}
      \includegraphics{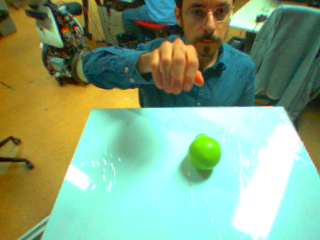}
      \includegraphics{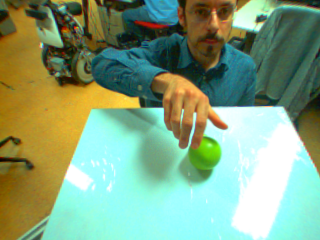}
      \includegraphics{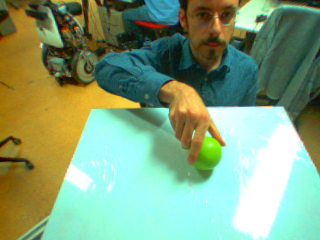}
      \includegraphics{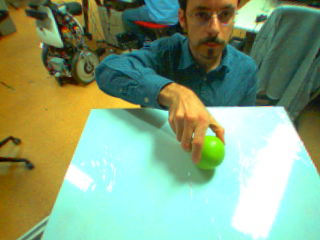}
      \includegraphics{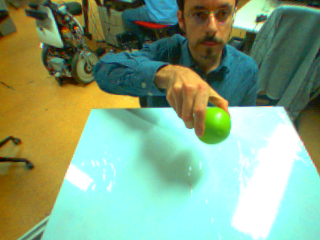}
    } % end resizebox
    \label{fig:descriptions:graspSphereGreenTwo}
  } % end subfloat

  \subfloat[][\touchSphereGreenTwo.]{
    \resizebox{\linewidth}{!}{
      \includegraphics{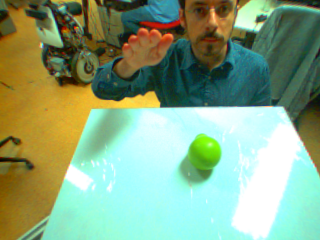}
      \includegraphics{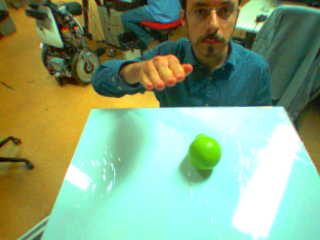}
      \includegraphics{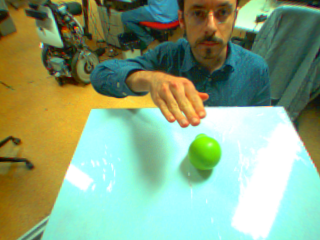}
      \includegraphics{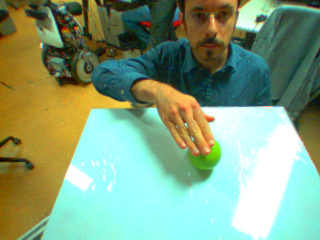}
      \includegraphics{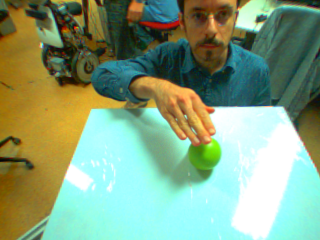}
      \includegraphics{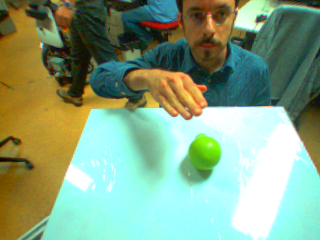}
    } % end resizebox
    \label{fig:descriptions:touchSphereGreenTwo}
  } % end subfloat

  \caption{Example of descriptions generated by the model.}
  \label{fig:descriptions}
\end{figure}

\subsection{Verbal Descriptions and Choice of Synonyms}
\label{sec:results:verbal_descriptions}

By generating and scoring natural language descriptions of what the robot observes~(see Sec.~\ref{sec:method:verbal}), we can provide evidence to the model and interpret the verbal results.
Recall that, with our method, we do not add new words to the model when we observe the human performing actions.
Rather, the human-readable descriptions that we generate are based on the same words that were present in the self-centered learning phase.
In this phase, the verbal descriptions described the agent of the observed actions is either ``the~robot'', ``he'', or ``Baltazar''~(the name of the robot).
Consequently, the \AffWords{} model learned by the robot includes those words as the subject of the action.

As an example, by providing the evidence \{Color=yellow, Size=big, Shape=sphere, ObjVel=fast\} to the model, we obtain the sentences reported in Table~\ref{tab:example_generated_sentences}.
The higher the score, the better.
In many of these sentences, we note that (i)~the correct verb related to the tap action is generated (in the initial evidence, no action information was present, only object features and effects information were), and (ii)~the object term ``ball'' or synonyms thereof~(e.g., ``sphere'') are used coherently, both in the first part of the sentence describing the action and in the second part describing the effect.
The fact that different synonyms may be used in the same sentence is simply a consequence of the random generation of sentences, described in Sec.~\ref{sec:method:verbal}, and of the fact that usually synonyms are assigned similar (but not necessarily equal) probabilities by the model, given the same evidence.

\subsection{Language Phenomenon: Choice of Correct Conjunction}
\label{sec:results:conjunction}

The manipulation experiments that we consider have the following structure: an agent~(human or robot) performs a physical action onto an object with certain properties, and this object will produce a certain physical effect as a result.
For example, a touch action on an object yields no physical movement, but a tap does~(especially if the object is spherical).
In the language description associated to an experiment, it makes sense to measure the conjunction chosen by the model given specific evidence.
In particular, it would be desirable to separate two kinds of behaviors: one in which the action and effect are coherent~(expected conjunction: ``and''), and the other one in which they are contradictory (``but'').

Fig.~\ref{tab:conjunction} shows an example of this behavior of the model.
We give the same action value \emph{grasp} to the model as evidence, but two different values for the final object velocity.
When the object velocity is medium~(Fig.~\ref{tab:conjunction:and}), the model interprets this as a successful grasp and uses the conjunction ``and'' to separate the description of the action from the description of the effect.
When the object velocity is slow~(in the clustering procedure, the velocity was most often zero in those cases), the model predicts that this is an unsuccessful grasp and uses the conjunction ``but'', instead.

\subsection{Language Phenomenon: Description of Object Features}
\label{sec:results:description_objects}

In Fig.~\ref{fig:descriptions}, we show examples of verbal descriptions generated by the model given different values of observed evidence:
\begin{itemize}
\item \small \evidenceProducingGraspBoxGreenOne \normalsize (\ref{fig:descriptions:graspBoxGreenOne});

\item \small \evidenceProducingTouchBoxGreenOne \normalsize (\ref{fig:descriptions:touchBoxGreenOne});

\item \small \evidenceProducingGraspSphereGreenTwo \normalsize (\ref{fig:descriptions:graspSphereGreenTwo});

\item \small \evidenceProducingTouchSphereGreenTwo \normalsize (\ref{fig:descriptions:touchSphereGreenTwo}).
\end{itemize}
Note that the box object in the two first examples has a dark shade of green~(value of Color affordance variable of Table~\ref{tab:bnsymb} clustered as:~green1), whereas the spherical one in the two last examples has a lighter shade~(Color value:~green2).
However, the verbal descriptions reported in Fig.~\ref{fig:descriptions} all use the adjective ``green''.
This behavior emerges from fact that the robot develops its perceptual symbols (clusters) in an early phase, and only subsequently associates them with the human vocabulary.
We believe that this phenomenon is practical and potentially useful~(i.e., the possibility that a low-level fine-grained robot representation can be abstracted into a high-level language description, which bundles the two shades of green under the same word).

%%%%%%%%%%%%%%%%%%%%%%%%%%%%%%%%%%%%%%%%%%%%%%%%%%%%%%%%%%%%%%%%%%%%%%%%%%%%%%%%
%!TEX encoding = UTF-8 Unicode

\section{Conclusions and Future Work}
\label{sec:conclusions}

We presented a model that allows a robot to interpret and describe the actions of external agents, by reusing the knowledge previously acquired in an ego-centric manner.
In a developmental setting, the robot first learns the link between words and object affordances by exploring its environment.
Then, it uses this information to learn to classify the gestures and actions of another agent.
Finally, by fusing the information from the two probabilistic models, in our experiments we show that the robot can reason over affordances and words when observing the other agent; this can also be leveraged to do early action recognition~(see Sec.~\ref{sec:results:anticipation_effects}).
Although the complete model only estimates probabilities of single words given the evidence, we showed that feeding these probabilities into a pre-defined grammar produces human-interpretable sentences that correctly describe the situation.
We also highlighted some interesting language-related properties of the model, such as:
congruent/incongruent conjunctions,
choice of appropriate synonym words,
describing object features with general words.

Our demonstrations are based on a restricted scenario~(see Sec.~\ref{sec:experimental_settings}), i.e., one human and one robot manipulating simple objects on a shared table, a pre-defined number of motor actions and effects, and a vocabulary of approximately~$50$ words to describe the experiments verbally.
However, one of the main strengths of our study is that it spans different fields such as robot learning, language grounding, and object affordances.
We also work with real robotic data, as opposed to learning images-to-text mappings~(as in many works in computer vision) or using robot simulations~(as in many works in robotics).

In terms of \emph{scalability}, note that our \ac{BN} model can learn both the dependency scructure and the parameters of the model from observations.
The method that estimates the dependency structure, in particular, is sensitive to biases in the data.
Consequently, in order to avoid misconceptions, the robot needs to explore any possible situation that may occur.
For example, if the robot only observes blue spheres rolling, it might infer that it is the color that makes the object roll, rather than its shape.
In order to scale the method to a larger number of concepts, it would be necessary to scale the amount of data considerably, similarly to what is typically done in deep learning.
In models of developmental robotics, where this is neither practically feasible, nor desirable, we would need to devise methods that can generalize more efficiently from very few observations.

As future work, it would be useful to investigate how the model can extract syntactic information from the observed data autonomously, thus relaxing the bag-of-words assumption in the current model.
Another line of research would be to study how the model can guide the discovery of new acoustic patterns~(e.g., \cite{falstrom:2017:glu, vanhainen:2014:icassp, vanhainen:2012:interspeech}), and how to incorporate the newly discovered symbols into our \AffWords{} model.
This would release our current assumption of a pre-defined set of words.

\appendices % for IEEEtran - https://tex.stackexchange.com/q/347621

%%%%%%%%%%%%%%%%%%%%%%%%%%%%%%%%%%%%%%%%%%%%%%%%%%%%%%%%%%%%%%%%%%%%%%%%%%%%%%%%
%!TEX encoding = UTF-8 Unicode

\section{Grammar Definition}
\label{appendix:grammar}
Below, we provide the grammar definition used to generate verbal descriptions from the probability distribution over words estimated by the model.
Note, however, that no grammar was used during the learning phase: the speech recognizer used as a frontend to the spoken descriptions is based on a loop of words with no grammar, and the \AffWords{} model is based on a bag-of-words assumption, where only the presence or absence of each word in the description is considered.
The symbol \texttt{.|.} represents alternative items, while the symbol \texttt{[.]} optional items.
Non-terminal symbols are given between \texttt{<.>} in italics, while words~(terminal symbols) are given in plain text and font: thus, the full set of words is given by all the plain text words below.

\begin{grammar}
  <sentence> ::= <agent> <action> <object> <conjunction> <object> <effect>

  <agent> ::= the robot | he | baltazar

  <action> ::= <touch> | <poke> | <tap> | <push> | <grasp> | <pick>

  <touch> ::= touches | [has] [just] touched | is touching

  <poke> ::= pokes | [has] [just] poked | is poking

  <tap> ::= taps | [has] [just] tapped | is tapping

  <push> ::= pushes | [has] [just] pushed | is pushing

  <grasp> ::= grasps | [has] [just] grasped | is grasping

  <pick> ::= picks | [has] [just] picked | is picking

  <object> ::= the [<size>] [<color>] <shape>

  <size> ::= big | small

  <color> ::= green | yellow | blue

  <shape> ::= sphere | ball | cube | box | square

  <conjunction> ::= and | but

  <effect> ::= <inertmove> | <slideroll> | <fallrise>

  <inertmove> ::= is inert | is still | moves | is moving

  <slideroll> ::= slides | is sliding | rolls | is rolling

  <fallrise> ::= rises | is rising | falls | is falling
\end{grammar}

%%%%%%%%%%%%%%%%%%%%%%%%%%%%%%%%%%%%%%%%%%%%%%%%%%%%%%%%%%%%%%%%%%%%%%%%%%%%%%%%
\printbibliography

%%%%%%%%%%%%%%%%%%%%%%%%%%%%%%%%%%%%%%%%%%%%%%%%%%%%%%%%%%%%%%%%%%%%%%%%%%%%%%%%
%!TEX encoding = UTF-8 Unicode

\begin{IEEEbiography}[{\includegraphics[width=1in,height=1.25in,clip,keepaspectratio]{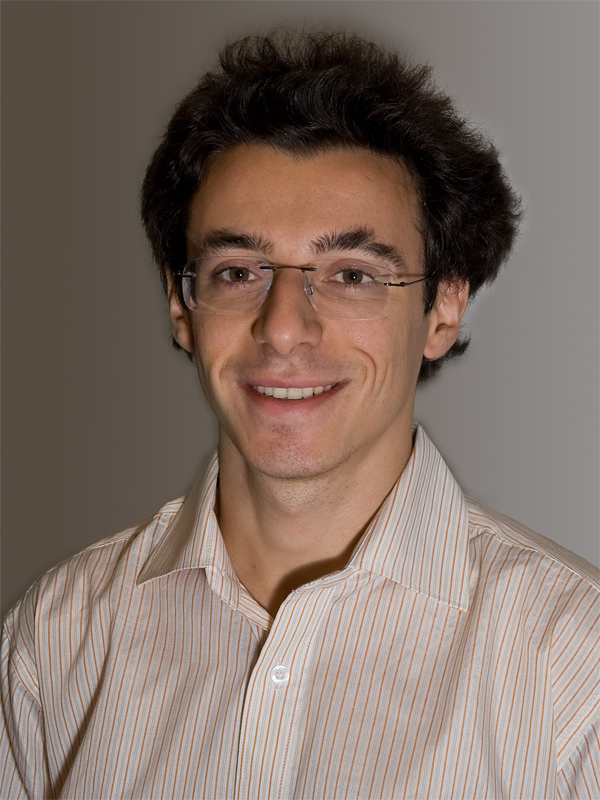}}]{Giovanni Saponaro}
  is a Ph.D. candidate at Instituto Superior Técnico (IST), Lisbon, Portugal, and a member of the local Computer and Robot Vision Laboratory (VisLab) of the Institute for Systems and Robotics. His research focuses on visual scene understanding and robot decision algorithms that support \hri{} with highly advanced humanoid robots such as the iCub, and in the presence of uncertainty. He received his M.Sc. in Computer Engineering -- \acl{AI} Systems from Sapienza University of Rome (Italy) in 2009 (with honors), and a B.Sc. in Computer Engineering from the same university in 2005. He published more than 10 papers in the diverse areas of cognitive systems, developmental robotics, visual perception of objects and of human body gestures for action recognition. He participated in the international research project POETICON++, together with linguists, computer vision experts, neuroscientists and roboticists.
\end{IEEEbiography}

\begin{IEEEbiography}[{\includegraphics[width=1in,height=1.25in,clip,keepaspectratio]{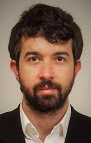}}]{Lorenzo Jamone}
  is a Lecturer in Robotics at the Queen Mary University of London (UK). He received his MS in Computer Engineering from the University of Genoa (Italy) in 2006 (with honors), and his PhD in Humanoid Technologies from the University of Genoa and the Italian Institute of Technology (IIT) in 2010. He was Associate Researcher at the Takanishi Laboratory in Waseda University (Tokyo, Japan) from 2010 to 2012, and Associate Researcher at VisLab laboratory of the Instituto Superior Tecnico (Lisbon, Portugal) from 2012 to 2016. His research interests include cognitive humanoid robots, sensorimotor learning and control, robotic manipulation, force and tactile sensing (more than 60 publications, h-index 16).
\end{IEEEbiography}

\begin{IEEEbiography}[{\includegraphics[width=1in,height=1.25in,clip,keepaspectratio]{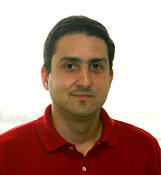}}]{Alexandre Bernardino}
  (PhD 2004) is an Associate Professor at the Dept. of Electrical and Computer Engineering and Senior Researcher at the Computer and Robot Vision Laboratory of the Institute for Systems and Robotics at IST, the faculty of engineering of Lisbon University. He has participated in several national and international research projects as principal investigator and technical manager. He published more than 40 research papers in peer-reviewed journals and more than 100 papers on peer-reviewed conferences in the field of robotics, vision and cognitive systems. He is associate editor of the journal Frontiers in Robotics and AI and of major robotics conferences (ICRA, IROS). He has graduated 10 PhD students and more than 40 MSc students. He was co-supervisor of the PhD Thesis that won the IBM Prize 2014 and the supervisor of the Best Robotics Portuguese MSc thesis award of 2012. He is a Senior Member of the IEEE and the current chair or the IEEE Portugal Robotics and Automation Chapter. His main research interests focus on the application of computer vision, machine learning, cognitive science and control theory to advanced robotics and automation systems.
\end{IEEEbiography}

\begin{IEEEbiography}[{\includegraphics[width=1in,height=1.25in,clip,keepaspectratio]{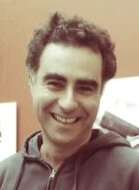}}]{Giampiero Salvi}
received the MSc degree in Electrical Engineering from Università la Sapienza (Rome, Italy) and the PhD degree in Computer Science from KTH Royal Institute of Technology (Stockholm, Sweden). He was a post-doctoral fellow at the Institute of Systems and Robotics (ISR), Lisbon, Portugal. He is currently Associate Professor in Machine Learning and Director of the Masters Programme in Machine Learning at KTH Royal Institute of Technology. His main interests are machine learning, speech technology, and cognitive systems.
\end{IEEEbiography}

\end{document}